\documentclass[preprint,11pt,numbers,sort&compress]{elsarticle}

\usepackage{amssymb,amsmath}
\usepackage{algorithm}
\usepackage{algorithmic}
\usepackage{multirow}
\usepackage{appendix}
\usepackage{changepage}
\usepackage{float}
\usepackage{caption}
\usepackage{bm}
\usepackage{threeparttable}
\usepackage{graphicx}
\usepackage{subfigure}

\usepackage{lineno}
\usepackage{url}

\usepackage{amsmath}
\usepackage{amssymb}
\usepackage{bm}
\DeclareMathOperator*{\argmax}{argmax}
\DeclareMathOperator*{\argmin}{argmin}
\usepackage{dashrule}
\usepackage{setspace}
\usepackage{bibspacing}
\usepackage[margin=3cm]{geometry}

\journal{Pervasive and Mobile Computing}

\renewcommand{\figurename}{Fig.}
\begin{document}

\begin{frontmatter}



\title{Learning Multi-level Features For Sensor-based Human Action Recognition}


\author[sklsdebhu,msra]{Yan Xu}
\ead{xuyan04@gmail.com}

\author[sklsdebhu]{Zhengyang Shen}
\ead{hbgtjxzbbx@gmail.com}
\author[sklsdebhu]{Xin Zhang}
\ead{xinzhang0376@gmail.com}
\author[sklsdebee]{Yifan Gao}
\ead{goghivan0017@gmail.com}
\author[sklsdebhu]{Shujian Deng}
\ead{cxwldsj@gmail.com}
\author[sklsdebhu]{Yipei Wang}
\ead{edithwang525@gmail.com}
\author[sklsdebhu]{Yubo Fan\corref{cor1}}
\ead{yubofan@buaa.edu.cn}
\cortext[cor1]{Corresponding author}
\author[msra]{Eric I-Chao Chang\corref{cor1}}
\ead{echang@microsoft.com}

\address[sklsdebhu]{State Key Laboratory of Software Development Environment and Key Laboratory of Biomechanics and Mechanobiology of Ministry of Education and Research Institute of Beihang University in Shenzhen, Beihang University, Beijing 100191, China}
\address[msra]{Microsoft Research Asia, Beijing 100080, China}
\address[sklsdebee]{School of Electronic and Information Engineering, Beihang University, Beijing 100191, China}

	\begin{abstract}
		This paper proposes a multi-level feature learning framework for human action recognition using a single body-worn inertial sensor. The framework consists of three phases, respectively designed to analyze signal-based (low-level), components (mid-level) and semantic (high-level) information. Low-level features capture the time and frequency domain property while mid-level representations learn the composition of the action. The Max-margin Latent Pattern Learning (MLPL) method is proposed to learn high-level semantic descriptions of latent action patterns as the output of our framework. The proposed method achieves the state-of-the-art performances, 88.7\%, 98.8\% and 72.6\% (weighted $F_1$ score) respectively, on Skoda, WISDM and OPP datasets.
	\end{abstract}

	\begin{keyword}
Multi-level \sep Human action recognition\sep Latent pattern \sep High-level  \sep Semantic
	

	\end{keyword}

\end{frontmatter}

\section{Introduction}
In recent years, sensor-based human action recognition (HAR) plays a key role in the area of ubiquitous computing due to its wide application in daily life \cite{chennuru2010mobile,wu2013mobisens,wu2011senscare,forster2009unsupervised,ngo2015similar,santos2015trajectory}. In most cases, utilization of raw data without special process is impractical since raw recordings confound noise and meaningless components. Therefore, the process of feature extraction and feature learning is supposed to be conducted \cite{lara2013survey}, which is commonly based on the sliding window technique through sampling overlapping frames from signal streams \cite{bulling2014tutorial}. In general, studies on sensor-based HAR have mainly focused on low-level \cite{bao2004activity,ravi2005activity,huynh2005analyzing,kang1995application} and mid-level \cite{bhattacharya2014using,vollmer2013learning,huỳnh2007scalable} features.

Low-level features include statistical features \cite{bao2004activity,ravi2005activity}, frequency-domain features \cite{huynh2005analyzing,kang1995application} and some hand-crafted methods \cite{hammerla2013preserving}. Low-level features are popular owing to their simplicity as well as their acceptable performances across a variety of action recognition problems. However, their simplicity is often accompanied by less discrimination in representation, considering that actions are often highly complex and diverse in daily-life cases.

Compared with low-level features, mid-level features which are mainly obtained through dictionary learning methods \cite{vollmer2013learning, huỳnh2007scalable} have proven to be more robust and discriminative \cite{bhattacharya2014using}. The representations include the sparse coding method \cite{grosse2012shift} and the bag-of-words (BOW) algorithm \cite{huỳnh2007scalable}. These methods analyze components (motion primitives) and explore inherent structures of signals. However, when the dictionary size is relatively large, mid-level features would suffer high computation, redundancy in representation and further burden the following classification. In this paper, we reduce the redundancy of the mid-level representation by introducing high-level features, which can achieve better overall performances and be robust to dictionary size.

Many studies have applied probabilistic graphical models \cite{zhang2013modeling,bui2008hidden} and pattern-mining methods \cite{liu2015action,chikhaoui2011frequent,gu2010unsupervised,kim2010human,palmes2010object} in semantic (high-level) recognition tasks. However, few works focus on feature learning methods in sensor-based HAR. In this paper, a semantic feature learning algorithm is proposed on two motivations. First, it can remove the redundancy in mid-level representation. A compact and discriminative description can be achieved by applying a latent pattern learning method on mid-level features. Second, implementing semantic analysis on sensor signals is an efficient and intuitive way to discover and deal with the variety of action patterns, and thus it can remove ambiguities in descriptions of the action. In particular, the ambiguities mainly come from two aspects, physical level and annotation level, both of which make it difficult to generate generalized good features in HAR tasks \cite{bulling2014tutorial}. At the physical level, one challenge is that different or even the same people who conduct the same action may produce completely different signals due to changes in environment, which is called the intra-class variability \cite{zinnen2009multi}. For example, walking style in morning after a good sleep commonly differs from the one after day's hard work. The other challenge is the inter-class similarity \cite{amft2007probabilistic}, which refers to the similarity among different actions such as `drinking milk' and `drinking coffee'. As the inter-class similarity is generally resolved through obtaining additional data from different sensors \cite{stikic2008adl} or analyzing co-occurring actions \cite{huynh2008discovery}, our work focuses on dealing with the intra-class variability. At the annotation level, a specific action performed in many ways can be cognitively identified as the same one \cite{aggarwal2011human}. Take `cleaning the table' as an example. It is rational to consider cleaning table from right to left and from up to down as the same action, though they behave absolutely differently in signal records. For those reasons, generalized features that are directly learned from action perspectives would compromise to the common characteristic shared by different action patterns, resulting in ambiguous feature representations \cite{malisiewicz2011ensemble}.
Inspired by Multiple-Instance Learning (MIL) \cite{dietterich1997solving,wang2013max}, our solution is to mine discriminative latent patterns from actions and construct features based on descriptions of those patterns, which can eliminate ambiguities from both physical and annotation levels. We name this method the Max-margin Latent Pattern Learning (MLPL) method. Instead of being constrained by generic property, the MIL method is a natural solution given that the diversity inside the class can be learnt. Although MIL methods are widely performed in computer vision \cite{ziaeefard2015semantic,yi2015human}, as for sensor-based HAR problems, relevant works are mainly designed to cope with sparse annotation \cite{stikic2011weakly,stikic2009activity}. Instead of dealing with sparsely labeled cases, MLPL proposed in this paper implements MIL to learn discriminative latent patterns of actions, by which high-level features would be acquired.

In this paper, we integrate the advantages of low-, mid- and high-level features and propose the framework known as Multi-Level Complementary Feature Learning (MLCFL). To avoid being confused with the high-level feature learned by the latent pattern learning process, the output feature of our framework is denoted as Multi-Level Complementary Feature (MLCF). In particular, this framework learns multi-level features through three phases, which are respectively designed to analyze signal-based (low-level), components (mid-level) and semantic (high-level) information. In the first phase, the low-level feature (statistical values, FFT coefficients, etc.) is extracted from raw signals. In the second phase, from the component perspective, the mid-level representation can be attained through hard coding processes and occurrence statistics. In the third phase, the MLPL method, from the semantic perspective, is implemented on the Compl feature (the concatenation of low- and mid-level features) to obtain MLCF as the output of this framework. Various experiments on Opp \cite{chavarriaga2013opportunity,roggen2010collecting}, Skoda \cite{zappi2008activity} and WISDM \cite{kwapisz2011activity} datasets show that MLCF possesses higher feature representation ability than low- and mid-level features. Moreover, compared with existing methods, the method we proposed achieves state-of-the-art performances. Our contributions in this paper are as follows:

\begin{enumerate}
	\item A multi-level feature learning framework MLCFL is constructed, which consists of three phases including low-level feature extraction, mid-level components learning and high-level semantic understanding. The output feature is learned level by level, possessing higher representation ability than low- and mid-level features.
	\item An efficient and effective latent pattern learning method MLPL is proposed to learn high-level features, each dimension of which refers to the confidence score of corresponding latent action pattern.
	\item Our framework is evaluated on three popular datasets and achieves state-of-the-art performances.
\end{enumerate}

The rest of this paper is organized as follows: Section 2 presents related work; Section 3 describes the MLCFL framework for action recognition; Section 4 presents and analyzes experimental results; finally, we conclude the study in Section 5.

\section{Related work}

Researches in the area of sensor-based HAR have been ever increasing in the past few years \cite{hammerla2013preserving,field2015recognizing,zhang2013modeling,bulling2014tutorial}. Numerous methods have been proposed in designing, implementing and evaluating action recognition systems. In this section, we review typically practical methods in terms of feature representation, specifically from low- \cite{bao2004activity,ravi2005activity,huynh2005analyzing, kang1995application,hammerla2013preserving,plotz2011feature} as well as mid- \cite{bhattacharya2014using,vollmer2013learning,zhang2012motion} perspectives and demonstrate how each method is applied to the specific action recognition task. As few works focus on designing high-level features, an additional review is presented on prevalent approaches involving semantic understanding of actions \cite{huynh2008discovery,field2015recognizing,zhang2013modeling,bui2008hidden,hospedales2011identifying,chikhaoui2011frequent,gu2010unsupervised,kim2010human,palmes2010object}. Besides, a brief overview about other representative methods \cite{stikic2011weakly,zeng2014convolutional} is summarized.

Low-level features are designed to capture signal-based information. Statistical metrics are the most common approaches, which include mean, variance, standard deviation, energy, entropy and correlation coefficients \cite{bao2004activity,ravi2005activity}. Fourier Transform (FT), Wavelet Transform (WT) \cite{tamura1997classification}, Discrete Cosine Transform (DCT) \cite{he2009activity} as well as auto-regressive (AR) coefficients \cite{he2008activity} are also commonly applied in HAR tasks for their promising performances. Kang et al. \cite{kang1995application} analyzed electromyography (EMG) signals by extracting conventional auto-regressive coefficients and cepstral coefficients as features. Hammerla et al. \cite{hammerla2013preserving} designed the hand-crafted feature based on the Empirical Cumulative Distribution Function (ECDF) to preserve characteristics of inertial signal distribution. Pl{\"o}tz et al. \cite{plotz2011feature} improved on that work and proposed the ECDF-PCA feature. They implemented the Principal Component Analysis (PCA) method on signals normalized by ECDF and significantly improved performance. In this paper, statistical values, FFT coefficients and ECDF-PCA are calculated as low-level features to demonstrate the generalization ability of the proposed framework.

Mid-level features are generally extracted from the low-level ones to explore the components and structural information of signals. They are prevalent in HAR tasks for robustness against noise and discrimination in representations \cite{bhattacharya2014using,vollmer2013learning}. Huỳnh et al. \cite{huỳnh2007scalable} and Zhang et al. \cite{zhang2012motion} implemented the bag-of-words (BOW) model to obtain statistical descriptions of motion primitives. Their works showed the effectiveness of the BOW model in sensor-based action recognition tasks. Blanke et al. \cite{blanke2009daily} extracted the occurrence statistics feature from low-level actions in a way that is similar to Huỳnh et al. \cite{huỳnh2007scalable} and then implemented the JointBoosting-framework. One characteristic of their method was to adopt a top-down perspective, using a feature selection algorithm to learn the distinctive motion primitives from the labeled high-level action. Sourav et al. \cite{bhattacharya2014using} and Christian et al. \cite{vollmer2013learning} both utilized sparse coding and adopted the convolution basis, which could resist shifts in time and thus reduce redundancy of basis. Our work is similar to \cite{zhang2012motion}, in which the mid-level representation is achieved through hard coding and occurrence statistics.

High-level recognition tasks mainly focus on obtaining intuitive and semantic descriptions of actions. Pattern-mining methods \cite{huynh2008discovery,chikhaoui2011frequent,gu2010unsupervised,kim2010human,palmes2010object,liu2015action} and probabilistic graphical models \cite{zhang2013modeling,bui2008hidden} are the most prevalent approaches. Pattern-mining methods explore the diversity in human actions through learning discriminative action patterns and motion primitives. Huynh et al. \cite{huynh2008discovery} applied probabilistic topic models which stemmed from the text processing community to automatically extract action patterns from sensor data. They described the recognition of daily routines as a probabilistic combination of learned patterns. Liu et al. \cite{liu2015action} presented an algorithm capable of identifying temporal patterns from low-level actions and utilized these temporal patterns to further represent high-level human actions. Various methods have also been proposed based on probabilistic graphical models \cite{zhang2013modeling,bui2008hidden} to capture the temporal, sequential, interleaved or concurrent relationship among motion primitives. However, graphical models are limited to capturing rich temporal relationships in complex actions and also suffer from the exponential increase in calculation when the number of involved actions grows \cite{liu2015action}. Instead of modeling temporal relationships, we propose an efficient pattern-mining approach, which takes advantage of being compact in representation, intuitive in understanding and efficient in calculation.

Multiple-Instance Learning (MIL) methods in HAR have been widely applied to cope with scarcity of annotation. Maja et al. \cite{stikic2011weakly} proposed a framework involving MIL as a weakly supervised recognition method to deal with scarcity of labels and proved its robustness to erroneous labels. Similarly, MIL methods with several novel extensions were introduced to handle different annotation strategies in action recognition \cite{stikic2009activity}. Instead of dealing with annotation scarcity, MLPL proposed in this paper implements MIL to explore latent patterns in human actions, by which the high-level feature would be acquired.

Other representative works in HAR include template matching methods, heuristic methods and deep learning ones. Template matching methods often derived from DTW and LCSS algorithm. Hartmann et al. \cite{hartmann2010gesture} proposed Segmented DTW to recognize and bound the gesture in each frame through finding the best match among the object and all templates of different classes. Nonsegmented DTW  proposed by  Stiefmeier et al. \cite{stiefmeier2008wearable} was a more efficient variation through reusing the previous computation. Nguyen-Din et al. \cite{nguyen2012improving,nguyen2014robust} improved LCSS algorithm and proposed WarpingLCSS and SegmentedLCSS, which were more efficient than DTW-based methods and robust to noisy annotation. Besides, heuristic methods are often related to specific tasks \cite{huynh2005analyzing}and depend on domain knowledge. Reyes-Ortiz et al. \cite{reyes2014human} designed temporal filters to recognize actions as well as postural transitions. In addition, various studies on deep learning methods have been conducted recently, mainly derived from Convolutional Neural Networks (CNN) \cite{ha2015multi,yang2015deep,ordonez2016deep} and Recurrent Neural Network (RNN) \cite{palumbo2016human}. These methods explored relationships between the temporal and spatial dependency of recordings and sensors. Zeng et al. \cite{zeng2014convolutional} applied CNN with partial weight sharing method. The framework they proposed achieved outstanding performance on three popular datasets.

In this paper, contrary to traditional feature extraction and learning methods restricted in low- and mid-level descriptions of signals, we achieve semantic understanding of sensor-based human actions through learning latent action patterns. To the best of our knowledge, our method proposed is the first attempt to explicitly apply a feature learning method to high-level representations in sensor-based HAR tasks. Furthermore, we present a brand new framework to synthesize multi-level features, integrating signal-based (low-level), components (mid-level) and semantic (high-level) information together.

\begin{figure}[H]
	\centering
	 \includegraphics[width=1\textwidth]{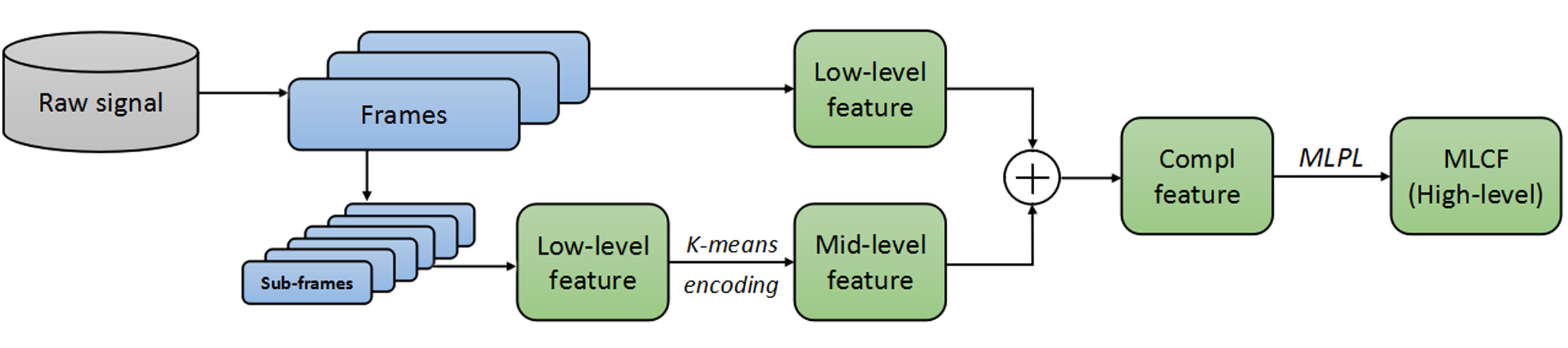}
	\caption{The flowchart of Multi-Level Complementary Feature Learning (MLCFL). In the first phase, low-level features are extracted from frames and sub-frames. In the second phase, mid-level representations can be obtained through hard coding processes and occurrence statistics. In the third phase, MLPL would be implemented on the Compl feature (the concatenation of low- and mid-level features), where MLCF can be obtained as the output of the framework.}
	\label{fig:overview}
\end{figure}

\section{Method}
Our framework consists of three phases in interpreting the signal: i) low-level feature extraction; ii) mid-level components learning; iii) high-level semantic understanding. The low-level description analyzes the temporal and frequency property of signals while the mid-level representation is a statistical description of either shared or distinctive components (motion primitives). The high-level feature describes the action by distinguishing the specific action pattern it belongs to. The flowchart of the framework is illustrated in \figurename\ \ref{fig:overview}.

\subsection{Low-level feature extraction}\label{section:3.1}
In this phase, low-level features are extracted from raw signals to learn properties in the time and frequency domain. Three popular features, namely statistical values, FFT coefficients and ECDF-PCA, are involved in this work and a brief description is presented as follows.

\subsubsection{Statistical values}
Statistical metrics analysis is one of the most common approaches to obtaining feature representations of raw signals. Statistical values in this work refer to time-domain features \cite{bao2004activity,ravi2005activity}, including mean, standard deviation, energy, entropy and correlation coefficients.

\subsubsection{FFT coefficients}
Frequency domain techniques \cite{huynh2005analyzing,kang1995application} have been extensively applied to capture the repetitive nature of sensor signals. This repetition often correlates to the periodicity property of a specific action such as walking or running. The transformation technique used in this paper is Fourier Transform (FT) through which dominant frequency components in the frequency domain can be procured.

\subsubsection{ECDF-PCA}
ECDF-PCA \cite{plotz2011feature} was brought up on the expertise analysis of inertial signals recorded by the triaxial accelerator sensor. It applies the Empirical Cumulative Distribution Function (ECDF) on signals. Then ECDF feature \cite{hammerla2013preserving} is procured by inverse equal probability interpolation, through which data are normalized and preserve its inherent structure at the same time. A PCA method is then implemented on this normalized data.

\subsection{Mid-level components learning}
The mid-level learning method is a general approach in pattern recognition. Dictionary learning methods such as bag-of-words (BOW) \cite{zhang2012motion} and sparse coding \cite{bhattacharya2014using} are the most popular approaches for obtaining mid-level representations. Compared with low-level feature extraction which involves analyzing properties in the time and frequency domain, mid-level learning focuses on the structural composition. In this paper, we implement BOW in signal processing. The obtained mid-level feature is the statistical description of components of the signal.

The dictionary is first formed from training data through the K-means algorithm \cite{huỳnh2007scalable}. In particular, frames are broken into overlapping sub-frames which are smaller in length. Low-level features are extracted from sub-frames and then the K-means method is used to construct the motion-primitive dictionary. $K$ clusters are generated and the dictionary is formed by cluster centers. We define a set of samples as $\{{\bm{\mathrm{x}}}_1,\dots,{\bm{\mathrm{x}}}_n\}, {\bm{\mathrm{x}}}_i\in \bm{\mathrm{R}}^{d\times1}$, $i\in\{1,\dots,n\}$ and each sample ${\bm{\mathrm{x}}}_i$ is associated with an index $z_i\in\{1,\dots,K\}$. If $z_i=j\in\{1,\dots,K\}$, ${\bm{\mathrm{x}}}_i$ is in the $j$-th cluster. The center of the $j$-th cluster is denoted as ${\bm{\mathrm{m}}}_j={\sum_{i=1}^{n}1\{z_i=j\}{\bm{\mathrm{x}}}_i}/{\sum_{i=1}^{n}1\{z_i=j\}}$, ${\bm{\mathrm{m}}}_j\in \bm{\mathrm{R}}^{d\times1}$. ${\bm{\mathrm{m}}}_j$ refers to a word while j refers to the corresponding index in the dictionary. When a new sample ${\bm{\mathrm{x}}}_i$ comes in, the corresponding index $z_i$ can be determined by $z_i=\mathop{\argmin}_{j\in\{1,\cdots,K\}}({\bm{\mathrm{x}}}_i-{\bm{\mathrm{m}}}_j)({\bm{\mathrm{x}}}_i-{\bm{\mathrm{m}}}_j)^\mathrm{T}$. 

\begin{figure}[H]
	\centering
	 \includegraphics[width=0.9\textwidth]{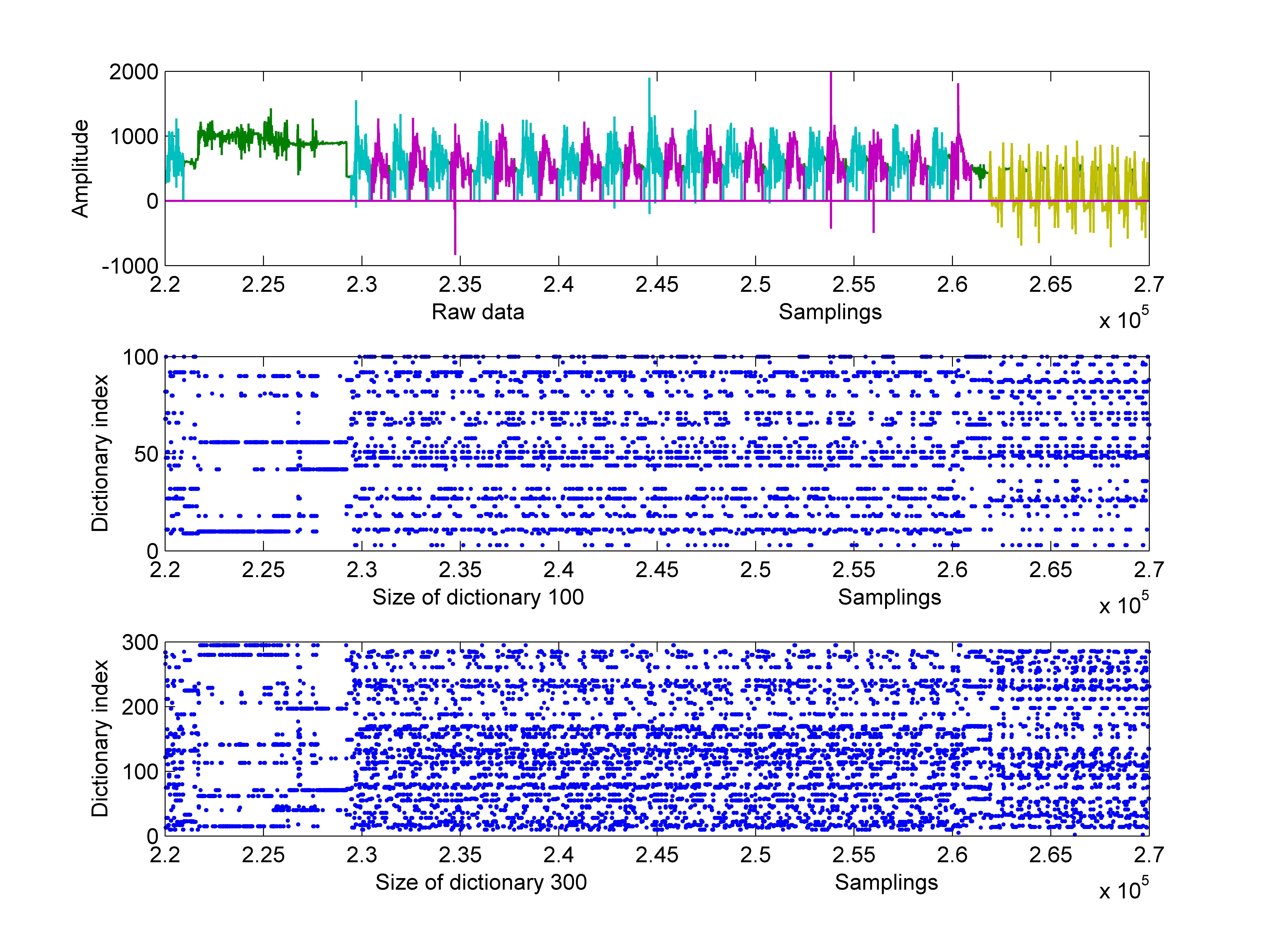}
	\caption{Illustration of symbolic sequences generated by different dictionary sizes in mid-level learning. From top to bottom: raw data, cluster assignments, of which the dictionary size is 100 and 300 respectively.}	
	\label{fig:fig2_symbolic_sequence}
\end{figure}

With the learned dictionary, a new symbolic sequence shown in \figurename\ \ref{fig:fig2_symbolic_sequence} can be derived from raw signals by densely extracting low-level features from sub-frames and retrieving them from the dictionary. In the end, the mid-level feature is represented by occurrence statistics of motion primitives in the frame.

\subsection{High-level semantic understanding}
\begin{figure}[H]
	\centering
	 \includegraphics[width=.5\textwidth]{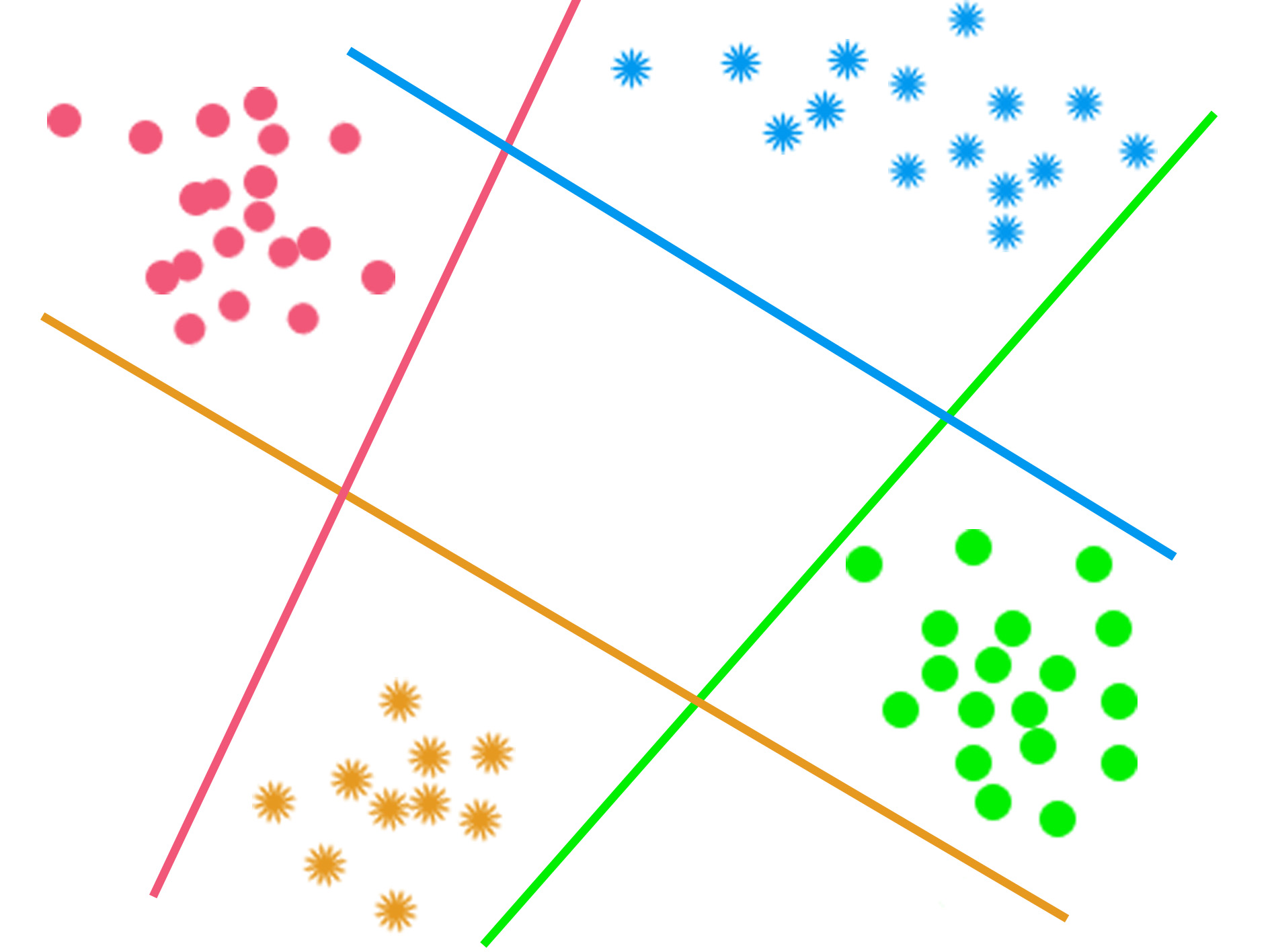}
	\caption{Illustration of Max-margin Latent Pattern Learning (MLPL). Circle and star represent two specific classes. Two different colors of each class are related to two latent classes which can be distinctive descriptions of action patterns. The confidence score is obtained by calculating the distance between the latent class margin and the instance in the feature space. The combination of the confidence scores of all the latent classes forms the semantic feature representation.}	
	\label{fig:MLPL}
\end{figure}

Generally speaking, one specific action would be highly identifiable if each pattern of the action had been learnt. Similarly, descriptions capturing properties of each distinctive distribution in the feature space would be more discriminative than the ones that only capture the generic but ambiguous characters of the whole distribution. Inspired by Wang's method \cite{wang2013max} of learning a weakly-supervised dictionary for discriminative subjects in images, we propose Max-margin Latent Pattern Learning (MLPL) in sensor-based signal processing. In general, the objective of this algorithm is to identify each specific class by learning a set of latent classifiers, each of which can be a discriminative description of a certain action pattern. The high-level feature is represented by the combination of confidence scores belonging to each latent class. Specifically, the latent pattern learning problem in this work, as shown in \figurename\ \ref{fig:MLPL}, can be divided into two aspects: i) maximizing the inter-class difference, namely the differences between one specific action and other actions; ii) maximizing the intra-class difference, namely the differences among different patterns of one specific action.

We first present a brief notation of Multiple-Instance Learning (MIL). In MIL, a set of bags are defined as {\bf{X}}=$\{{\bm{X}}_1,\dots,{\bm{X}}_m\}$, and each bag contains a set of instances ${\bm{X}}_i=\{{\bm{\mathrm{x}}}_{i1},\dots,{\bm{\mathrm{x}}}_{in}\}$, where ${\bm{\mathrm{x}}}_{ij}\in{{\bm{\mathrm{R}}}^{d\times1}}$. Only one label can be assigned to a bag and instances inside it. The bag would be labeled as positive if there exists at least one positive instance, while being labeled negative only when all of the instances in it are negative. In computer vision, the bag model is a natural description of the image because the image commonly consists of a set of subjects and the label of the image can only be determined by subjects of interest. Compared with the traditional MIL problem, the concept `bag' is simplified in our problem as labels of the whole signal are provided. Therefore, learning the `interest' from the background would be evaded. The goal of our method is to learn various patterns of actions.

\begin{figure}[H]
	\centering
	 \includegraphics[width=.9\textwidth]{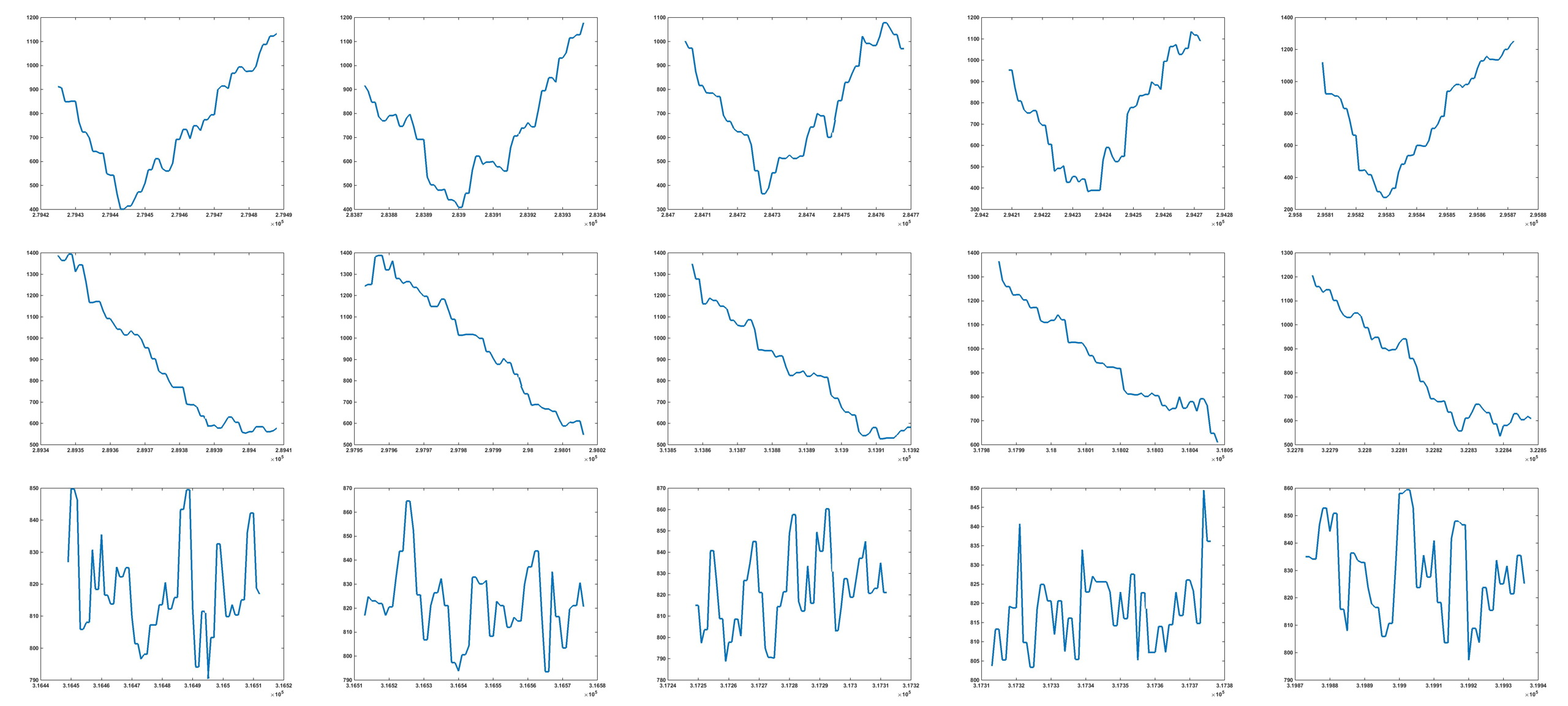}

      (a)

	 \includegraphics[width=.9\textwidth]{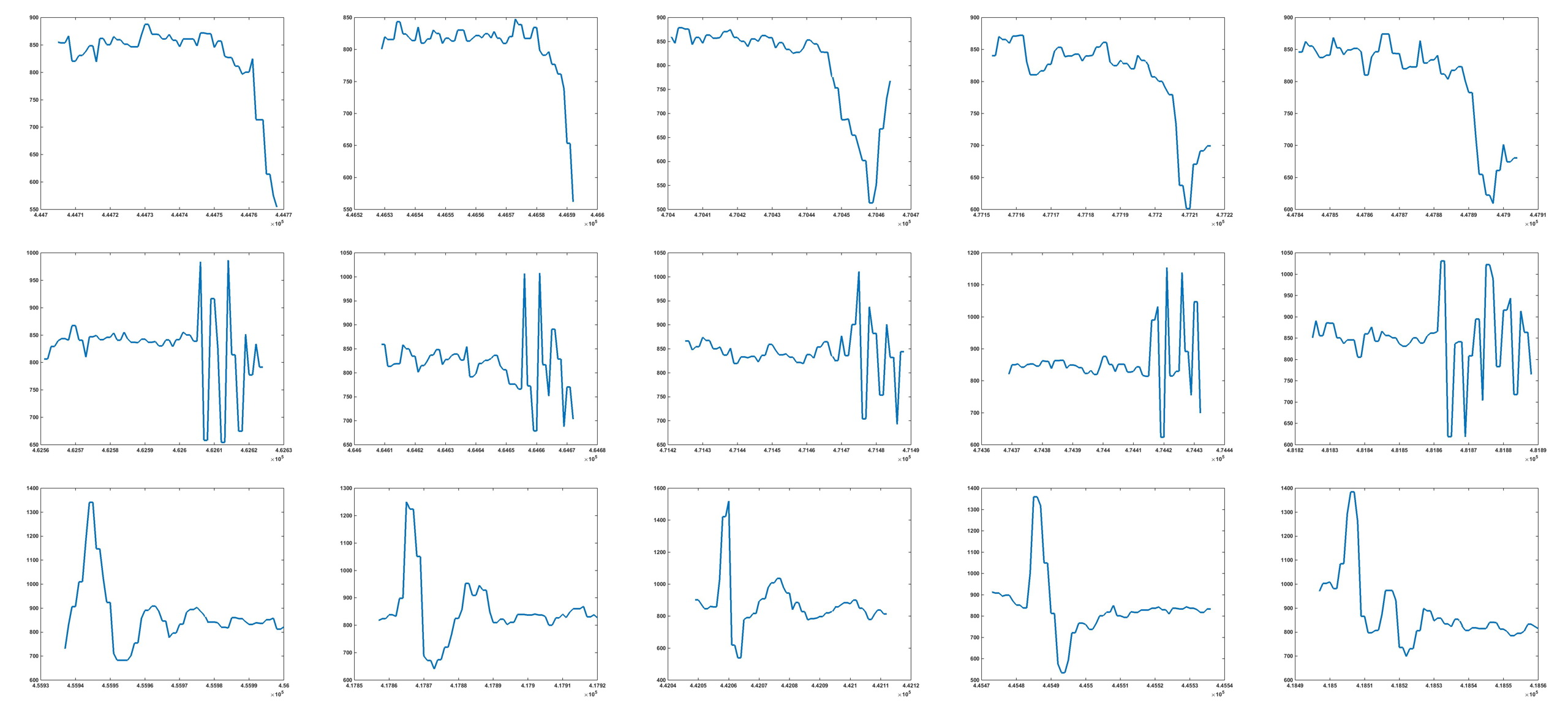}

      (b)
	\caption{Illustrations of latent class. Two classes in Skoda, `check gaps on the front door' (a) and `close both left door' (b), with their three latent classes are shown in raw signal. Each row refers to a latent class, where five samples are selected. Rows compare differences among latent classes and columns show the similarity in one specific latent class.}
	\label{fig:check steering wheel}
\end{figure}

To simplify the notation, we define instances from the $i$-th class, $i\in\{1,\dots,m\}$, as a set ${\bm{X}}_i=\{{\bm{\mathrm{x}}}_{i1},\dots,{\bm{\mathrm{x}}}_{in}\}$, where ${\bm{\mathrm{x}}}_{ij}\in{{\bm{\mathrm{R}}}^{d\times1}}$, $j\in\{1,\dots,n\}$. In addition, each set ${\bm{X}}_i$ is associated with the label $Y_i\in\{0,1\}$. When modeling latent patterns in the $i$-th class, $Y_i=1$, otherwise $Y_i=0$. For each class, we assume it contains $K$ latent classes, each of which corresponds to a cluster in the feature space. Intuitively, we assume there are 5 different patterns in class `check gaps on the front door' and `close both left door' respectively.  \figurename\ \ref{fig:check steering wheel} shows examples of patterns (latent classes) learned by MLPL. Multi-scale policy \cite{park2008hierarchical} is introduced in this paper to learn latent patterns from various semantic levels as well as mitigate uncertainty in determining concrete number of latent classes.

During the learning phase, each instance ${\bm{\mathrm{x}}}_{ij}$ is associated with a latent variable $z_{ij}\in\{0,1,\dots,K\}$. Instance ${\bm{\mathrm{x}}}_{ij}$ is in the $k$-th positive cluster if $z_{ij}=k$ or in negative cluster if $z_{ij}=0$. Considering that MIL is applied in our method to find distinctive latent classes, maximizing differences between one latent class and others is required. Besides, latent classes that belong to the $i$-th class need to be distinguished from the $j$-th class, where $i\neq{j}$. Consequently, from the feature distribution perspective, a natural idea is to maximize the margin among intra- and inter-class. To meet that demand, multi-class SVM is an ideal solution. In particular, SVM with linear kernel is used for its generality and efficiency. Each latent class is associated with a linear classifier, in which $f({\bm{\mathrm{x}}})={\bm{\mathrm{w}}}^\mathrm{T}{\bm{\mathrm{x}}}$ and weighting matrix is defined as
\begin{equation}
{\bm{\mathrm{W}}}_i=[{\bm{\mathrm{w}}}_0,{\bm{\mathrm{w}}}_1,\dots,{\bm{\mathrm{w}}}_K],{\bm{\mathrm{w}}}_k\in {{{\bm{\mathrm{R}}}^{d\times1}}},k\in{[0,1,\dots,K]}
\end{equation}
where ${\bm{\mathrm{w}}}_k$ represents the model of the $k$-th latent class if k is positive; ${\bm{\mathrm{w}}}_0$ denotes the negative cluster model. Therefore, we learned $K+1$ linear classifiers for each class and $m*(K+1)$ for total.

Intuitively, the latent label of the instance ${\bm{\mathrm{x}}}_{ij}$ is determined as the most `positive' one.
\begin{equation}
z_{ij} = \mathop{\argmax}_{k\in\{0,\cdots,K\}}{\bm{\mathrm{w}}}_k^\mathrm{T}{\bm{\mathrm{x}}}_{ij}
\end{equation}

The multi-class hinge loss forces latent classifiers to be distinct from each other. It can be defined as:
\begin{equation}
L({\bm{\mathrm{W}}};({\bm{\mathrm{x}}}_{ij},z_{ij}))=\sum_{ij} max(0,1+{\bm{\mathrm{w}}}_{r_{ij}}^\mathrm{T}{\bm{\mathrm{x}}}_{ij}-{\bm{\mathrm{w}}}_{z_{ij}}^\mathrm{T}{\bm{\mathrm{x}}}_{ij})
\end{equation}
where $r_{ij}=\mathop{\argmax}_{k\in\{0 \cdots K\},k\neq z_{ij}}{\bm{\mathrm{w}}}_{k}^\mathrm{T}{\bm{\mathrm{x}}}_{ij}$.

The objective function can then be defined as:
\begin{equation} \label{obj_eq}
\begin{split} 
\min_{{\bm{\mathrm{W}}},z_{ij}}\sum_{k=0}^{K}{\lVert{{\bm{\mathrm{w}}}_k}\rVert}^2+\alpha\sum_{ij}max(0,1+{\bm{\mathrm{w}}}_{r_{ij}}^\mathrm{T}{\bm{\mathrm{x}}}_{ij}-{\bm{\mathrm{w}}}_{z_{ij}}^\mathrm{T}{\bm{\mathrm{x}}}_{ij}),\\
s.t.\; if\; Y_i=1, z_{ij}>0,and\; if\; Y_i=0,z_{ij}=0,\\
\end{split}
\end{equation}
$where\; r_{ij}=\mathop{\argmax}_{k\in\{0,\cdots,K\},k\neq z_{ij}}{\bm{\mathrm{w}}}_k^\mathrm{T}{\bm{\mathrm{x}}}_{ij}.$

The first term in Eq. (\ref{obj_eq}) is for margin regularization and the second term is the multi-class hinge loss maximizing both inter- and intra-class margins. $\alpha$ balances the weight between two terms.
In MLPL, all negative instances are utilized in the optimization step. Latent labels in each class are initialized by K-means and updated according to their `positiveness' to each latent class. Constraint in Eq. (\ref{obj_eq}) forces the function to learn latent patterns in the $i$-th class. Though the optimization solution problem is a non-convex one, a local optimization can be guaranteed once the latent information is given \cite{felzenszwalb2010object}. We take ${\bm{\mathrm{w}}}_{k}^\mathrm{T}{\bm{\mathrm{x}}}_{ij}$ as confidence scores of ${\bm{\mathrm{x}}}_{ij}$ belonging to the $k$-th latent class. Output descriptors are thus represented by the combination of confidence scores belonging to each latent class. Our method is different from \cite{wang2013max} in three aspects. First, our work focuses on mining latent patterns rather than differentiating subjects of interests from the background. Second, latent classes are learned from all positive instances. The instance selection is removed as there are no background subjects in our problem. Third, instead of using the fixed number of latent classes, multi-scale strategy is adopted so that latent classes can be learnt at different semantic scales. We concatenate features learned by each scale.

\begin{algorithm}[H]
\caption{Max-margin Latent Pattern Learning: MLPL}
\label{mm-MCIL}
\begin{algorithmic}[!htbp]
\STATE {\bf\emph{{Learning latent patterns}}}
\STATE {\bf{Input}}: Training Instances {\bf{X}}=$\{{\bm{X}}_1,\dots,{\bm{X}}_m\}$, number of latent patterns per class K
\STATE {\bf{Loop $\ell$}}: Positive instances ${\bm{X}}_\ell$, negative instances {\bf{X}}$-{\bm{X}}_\ell$
\vspace{-0.6cm}
\begin{adjustwidth}{0.5cm}{0cm}
\STATE {\bf{Initialization}}: For negative instances, $z_{ij}$=0. For positive instances, $z_{ij}$ is initialized by K-means. 
   \STATE {\bf{Iteration}}: $N$ times 
\end{adjustwidth}
\vspace{-0.6cm}
\begin{adjustwidth}{1.2cm}{0cm}
    \STATE {\bf{Optimize ${\bm{\mathrm{W}}}_\ell$}}: Solve the multi-class SVM optimization problem
	\begin{eqnarray*}
	\min_{{\bm{\mathrm{W}}}_\ell}\sum_{k=0}^{K}{\lVert{{\bm{\mathrm{w}}}_k}\rVert}^2+\alpha\sum_{ij}max(0,1+{\bm{\mathrm{w}}}_{r_{ij}}^\mathrm{T}{\bm{\mathrm{x}}}_{ij}-{\bm{\mathrm{w}}}_{z_{ij}}^\mathrm{T}{\bm{\mathrm{x}}}_{ij}),	
	\end{eqnarray*}
	\quad \quad $where\; r_{ij}=\mathop{\argmax}_{k\in\{0 \cdots K\},k\neq z_{ij}}{\bm{\mathrm{w}}}_{k}^\mathrm{T}{\bm{\mathrm{x}}}_{ij}$.

	\STATE {\bf{Update $z_{ij}$}}: For positive instances:
	\begin{eqnarray*}
	z_{ij}=\mathop{\argmax}_{k\in\{1,\cdots,K\}}({\bm{\mathrm{w}}}_k^\mathrm{T}{\bm{\mathrm{x}}}_{ij}-{\bm{\mathrm{w}}}_0^\mathrm{T}{\bm{\mathrm{x}}}_{ij}),
	\end{eqnarray*}
\end{adjustwidth}
\STATE {\bf{Output}}: The learned classifiers {\bm{\mathrm{W}}}=[${\bm{W}}_1$,\dots,${\bm{W}}_m$].
~\\
\hdashrule[0.5ex]{12cm}{1pt}{3pt}
\STATE {\bf\emph{{Semantic representation}}}
\STATE {\bf{Input}}: Instances {$\bf{X}$}
\STATE {\bf{Output}}: $\bf{F(X)}=\bf{W}^\mathrm{T}\bf{X}$
\end{algorithmic}
\end{algorithm}

In our MLCFL framework, the MLPL process is implemented on the concatenation of low- and mid-level features, where we obtain MLCF as the output of the feature learning stage and input of the classification stage.

\section{Experiments and results}
In this section, we first describe the datasets, evaluation method and experimental settings for the framework. Moreover, we compare our framework, MLCFL, with other closely related methods and test the framework with three classifiers. Then we demonstrate the effectiveness of Max-margin Latent Pattern Learning (MLPL), the complementary property of low-level and mid-level features. We further conduct intra-personal (tasks performed by one person) and inter-personal (tasks performed by different persons) experiments. In the end of this section, we explore the sensitivity of parameters.

\subsection{Dataset}
We evaluate the proposed Multi-Level Complementary Feature Learning (MLCFL) framework on three popular datasets, namely Skoda, WISDM and Opp. Experimental settings on the three datasets are listed as follows.

\begin{adjustwidth}{0.5cm}{0cm}
\textbf{Skoda} \cite{zappi2008activity} Skoda Mini Checkpoint contains 16 manipulative gestures, which are performed in a car maintenance scenario and collected by 3D acceleration sensors. Only one worker's data are recorded, including actions such as `open hood', `close left hand door', etc. The sampling rate is 96Hz. In our experiments, 10 actions and a null class from one right arm-sensor are taken into consideration. 4-fold cross validation is conducted.
\end{adjustwidth}

\begin{adjustwidth}{0.5cm}{0cm}
\textbf{WISDM} \cite{kwapisz2011activity} The dataset is collected by Android-based smart phones in 36 users' pockets. Records comprise 6 daily actions such as `jogging', `walking', etc. Data are collected through controlled laboratory conditions and the sampling rate is 20Hz. We conduct 10-fold cross validation on this dataset.
\end{adjustwidth}

\begin{adjustwidth}{0.5cm}{0cm}
\textbf{Opp} \cite{chavarriaga2013opportunity,roggen2010collecting} Opportunity Activity Recognition is collected by 242 attributes in the scene that simulates a studio flat. The sampling rate of triaxial accelerometers is 64Hz. In this paper, only data collected by the single inertial sensor on the right low arm are utilized, including 13 low-level actions (`clean', `open' and `close', etc.) and a null class. 5-fold cross validation is conducted on this dataset.
\end{adjustwidth}

\subsection{Evaluation method} 
In HAR, data are severely unbalanced. Some classes are overrepresented while others are scarce. To adapt this characteristic, we apply the weighted $F_1$ score for evaluation.

\textbf{Weighted $F_1$ Score} We denote TP and FP as the number of true positives and false positives respectively, and FN as the number of false negatives. Thus, weighted $F_1$ score can be formulated as follows: 
\begin{equation}
precision_i=\frac{TP_i}{TP_i+FP_i}
\end{equation}
\begin{equation}
recall_i=\frac{TP_i}{TP_i+FN_i}
\end{equation}
\begin{equation}
F_1=\sum_i 2\cdot w_i\frac{precision_i\cdot{recall_i}}{precision_i+recall_i}
\end{equation}
in which $i$ refers to the class index and $w_i$ is the proportion of the $i$-th class in all the classes.
Unless mentioned otherwise, we adopt weighted $F_1$ score throughout experiments.

\subsection{Experiment setup} 

To compare with other methods, we follow the dataset settings in Zeng's work \cite{zeng2014convolutional}. Specifically, in all three datasets, we utilize data from a single triaxial inertial sensor. Sensor data are segmented into frames using a sliding window with the size of 64 continuous samples and with 50\% overlap. We also test window size of 48, 80 and get F1-value 86.8\%, 89.5\% on Skoda respectively. Since a frame may contain different labels, only frames with a single label are taken into consideration in Skoda and WISDM. But when it comes to Opp, to obtain enough samples, the label of the frame is determined according to the dominant label. Consequently, Skoda, WISDM and Opp contain around 22,000, 33,000 and 21,000 frames, respectively. In cross-validation, folds are created by randomly choosing samples from the dataset. 

To demonstrate the generalization ability of the proposed framework, we test three prevalent low-level features, namely statistical values, FFT coefficients and ECDF-PCA. We calculate statistical values (mean, standard deviation, energy, entropy and correlation coefficients) and FFT coefficients the same way as Pl{\"o}tz et al. \cite{plotz2011feature} and obtain 23-dimension and 30-dimension feature vectors for each frame respectively. In our practice, statistical values and FFT coefficients with Z-score normalization gain significant improvement in performances than original ones. For ECDF-PCA, we normalize the raw signal through 60 points inverse equal probability interpolation along each channel. Then the PCA process is conducted, where 30 principal components (30 dimensions) are taken as output features. In the mid-level learning phase, the length of the sub-frame for encoding process is set to 20, and the size of the dictionary is set to 300. We concatenate features from three channels of a single triaxial inertial sensor and yield 900-dimension sparse representations. In the high-level learning phase, the input feature is the concatenation of a low-level feature (statistical values with 23 dimensions, FFT coefficients with 30 dimensions, ECDF-PCA with 30 dimensions) and its corresponding mid-level feature (900 dimensions). The weight parameter $\alpha$ in Eq. (\ref{obj_eq}) is set to 1. Besides, the latent-class number $K$ for each class is set to 5 and 10 in two scales. In the step of optimizing ${\bm{\mathrm{W}}}$, we apply linear SVM \cite{fan2008liblinear} to handle multi-class classification problem. The number of iterations $N$ is set to 3. Without loss of generality, settings for features are the same throughout experiments. 

We test the performance of the MLCFL framework with three representative classifiers, namely K-Nearest Neighbors (KNN), Support Vector Machine (SVM) and Nearest Centroid Classifier (NCC). In the K-Nearest Neighbors algorithm (KNN), samples are classified by a majority vote of their neighbors and assigned to the most common label among its $K$ nearest neighbors. $K$ is set to 5 throughout our experiments. Besides, SVM with linear kernel is a prevalent and efficient solution in classification. Specifically, Liblinear \cite{fan2008liblinear} is used in our experiment\footnote{Different from the Liblinear in high-level feature learning, here we use it as a classifier}. Nearest Centroid Classifier (NCC) \cite{chavarriaga2013opportunity} calculates Euclidean Distance between test samples and each centroid (mean) of training classes. The predicted label is allocated according to the nearest class centroid. 

\subsection{Experiments} 
In the following sections, we conduct experiments in five respects: (1) comparing with existing methods; (2) conducting classifiers experiments on MLCF; (3) evaluating the effectiveness of MLPL; (4) exploring the complementary property of the low-level feature and mid-level feature; (5) performing intra- and inter-personal experiments.

\subsubsection{Comparing with existing methods}

In this section, comparisons of our framework with several published works are shown in Table \ref{table:compare with existing methods}, including representative methods: statistical values, FFT coefficients, ECDF-PCA \cite{plotz2011feature} and W-CNN \cite{zeng2014convolutional}. W-CNN applies CNN with the partial weight sharing technique to perform HAR tasks. This method achieves high performances in single body-sensor based classification tasks on three popular datasets. Statistical values, FFT coefficients and ECDF-PCA are respectively utilized as low-level features in our framework. Their corresponding output features are abbreviated to Stat-MLCF, FFT-MLCF and ECDF-PCA-MLCF. We follow the metrics in the W-CNN work and present results using KNN in the form of classification accuracy.

\begin{table*}[!htbp]
	\caption{\label{table:compare with existing methods} \newline Comparison (classification accuracy) of our presented MLCFL algorithm with existing methods on three datasets.}
	\centering
	\small
	\scalebox{1}{
	\begin{threeparttable}
	\begin{tabular}{ccccccc}
		\hline
		Feature & Skoda & WISDM & Opp\\
		\hline
		Statistical & 83.6 & 95.8 & 69.4\\
		FFT & 81.3 & 96.4 & 69.2\\
		ECDF-PCA \cite{plotz2011feature} & 85.4 & 98.1 & 69.4\\
		W-CNN \cite{zeng2014convolutional} & 88.2 & 96.9 & /\tnote{a}\\
		\hdashrule[0.5ex]{3cm}{1pt}{3pt} &\hdashrule[0.5ex]{2cm}{1pt}{3pt} &\hdashrule[0.5ex]{2cm}{1pt}{3pt} &\hdashrule[0.5ex]{2cm}{1pt}{3pt}\\
		\emph{Stat-MLCF} & 88.4 & 98.2 & 72.8\\
		\emph{FFT-MLCF} & 87.6 & 98.3 & \textbf{73.5}\\
		\emph{ECDF-PCA-MLCF} & \textbf{88.7} & \textbf{98.8} & 72.6\\
		\hline
	\end{tabular}
	\begin{tablenotes}
        \footnotesize
        \item[a] Performances of W-CNN on 13 actions are not accessible.
      \end{tablenotes}
     \end{threeparttable}}
\end{table*}

Table \ref{table:compare with existing methods} illustrates that the general performance of MLCF is the highest. Since our algorithm can be applied as the refining process of these low-level features, improvements on statistical values, FFT coefficients and ECDF-PCA are obvious. Despite parameters being generalized on all three datasets, results show that our method still achieves better results than the state-of-the-art method, W-CNN \cite{zeng2014convolutional}, on available Skoda and WISDM. This shows the effectiveness of MLCFL which integrates advantages of low-, mid- and high-level analyzes.

\begin{table*}[!htbp]
	\caption{\label{table:MLCF} \newline Comparison (weighted $F_1$ score) among low-level features, their mid-level features and MLCF on three datasets using three classifiers.} 
	\centering
	\small
	\scalebox{1}{\textbf{KNN}}\\
	\scalebox{0.8}{
	\begin{tabular}{cccccccccc}
		\hline
		\multirow{2}{*}{Feature} &
		\multicolumn{3}{c}{Skoda} & 
		\multicolumn{3}{c}{WISDM} & 
		\multicolumn{3}{c}{Opp}\\
		\cline{2-10}
		& low-level & mid-level & MLCF & low-level & mid-level & MLCF & low-level & mid-level & MLCF\\
		\hline
		Statistical & 83.6 & 85.3 & \textbf{88.5} & 95.6 & 96.9 & \textbf{98.2} & 67.9 & 68.5 & \textbf{70.8} \\
		FFT & 81.3 & 82.7 & \textbf{87.6} & 96.3 & 97.1 & \textbf{98.3} & 67.3 & 69.1 & \textbf{71.7}\\
		ECDF-PCA & 85.4 & 84.3 & \textbf{88.7} & 98.1 & 98.0 & \textbf{98.8} & 67.6 & 67.8 & \textbf{70.5}\\
		\hline
	\end{tabular}}
	\scalebox{1}{ }\\
	\scalebox{1}{\textbf{SVM}}

	\scalebox{0.8}{
	\begin{tabular}{cccccccccc}
		\hline
		\multirow{2}{*}{Feature} &
		\multicolumn{3}{c}{Skoda} & 
		\multicolumn{3}{c}{WISDM} & 
		\multicolumn{3}{c}{Opp}\\
		\cline{2-10}
		& low-level & mid-level & MLCF & low-level & mid-level & MLCF & low-level & mid-level & MLCF\\
		\hline
		Statistical & 54.9 & 81.8 & \textbf{84.4} & 73.5 & 92.3 & \textbf{93.9} & 53.5 & 63.7 & \textbf{66.2} \\
		FFT & 50.8 & 81.6 & \textbf{82.6} & 86.6 & 93.6 & \textbf{95.4} & 50.6 & 64.8 & \textbf{66.7}\\
		ECDF-PCA & 40.3 & 82.4 & \textbf{83.7} & 77.4 & 95.5 & \textbf{96.1} & 47.4 & 64.6 & \textbf{65.7}\\
		\hline
	\end{tabular}}
	\scalebox{1}{ }\\
	\scalebox{1}{\textbf{NCC}}\\

	\scalebox{0.8}{
	\begin{tabular}{cccccccccc}
		\hline
		\multirow{2}{*}{Feature} &
		\multicolumn{3}{c}{Skoda} & 
		\multicolumn{3}{c}{WISDM} & 
		\multicolumn{3}{c}{Opp}\\
		\cline{2-10}
		& low-level & mid-level & MLCF & low-level & mid-level & MLCF & low-level & mid-level & MLCF\\
		\hline
		Statistical & 49.6 & 64.4 & \textbf{65.4} & 73.3 & 83.1 & \textbf{83.3} & 45.8 & 48.4 & \textbf{51.5} \\
		FFT & 40.6 & 61.9 & \textbf{67.2} & 79.6 & 81.1 & \textbf{88.6} & 47.4 & 43.2 & \textbf{53.9}\\
		ECDF-PCA & 37.2 & 63.6 & \textbf{66.6} & 65.8 & 87.0 & \textbf{89.8} & 49.1 & 52.0 & \textbf{52.3}\\
		\hline
	\end{tabular}}
\end{table*}

\subsubsection{Results from different classifiers}

In this section, we demonstrate the `good' feature property of MLCF by presenting results from three classifiers. Results from all three datasets are shown in Table \ref{table:MLCF}.

It can be observed from Table \ref{table:MLCF} that MLCF performs better than low- and mid-level features in all three classifiers and achieves the best performance using KNN. Though the sparse representation is preferred by linear SVM, MLCF of 187 (Skoda), 102 (WISDM) and 238 (Opp) dimensions still show obvious enhancement compared with mid-level features of 900 dimensions. Besides, the decision process of NCC is based on the distribution of the feature, which suggests that the distribution of MLCF in the feature space is more regular and discriminative.

\subsubsection{The effectiveness of the MLPL process}
\begin{figure}[H]
	\centering

		\begin{minipage}[b]{.32\textwidth}
	 	 \includegraphics[width=1\textwidth]{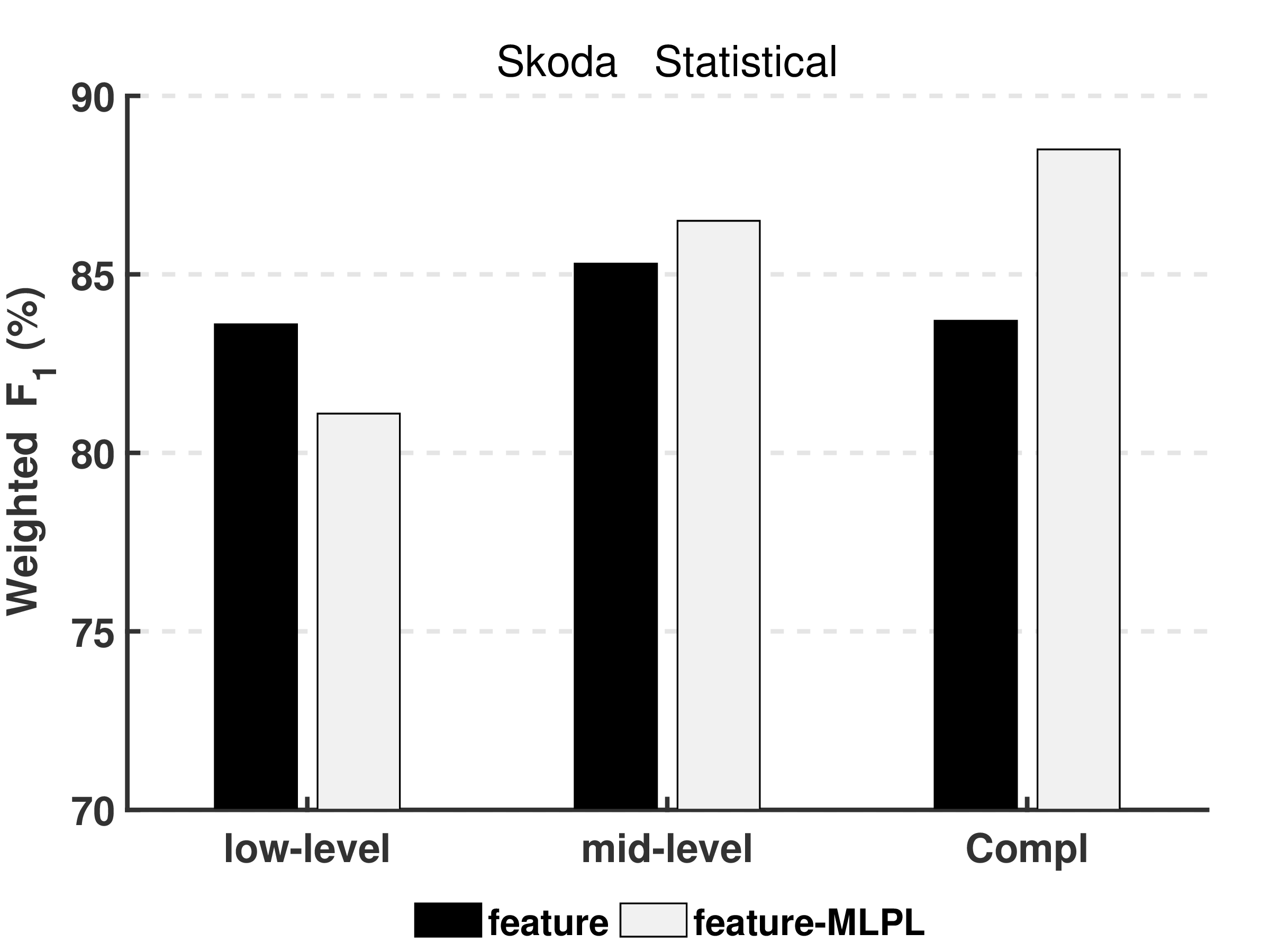}
		\end{minipage}
		\hfill
		\begin{minipage}[b]{.32\textwidth}   
	 	 \includegraphics[width=1\textwidth]{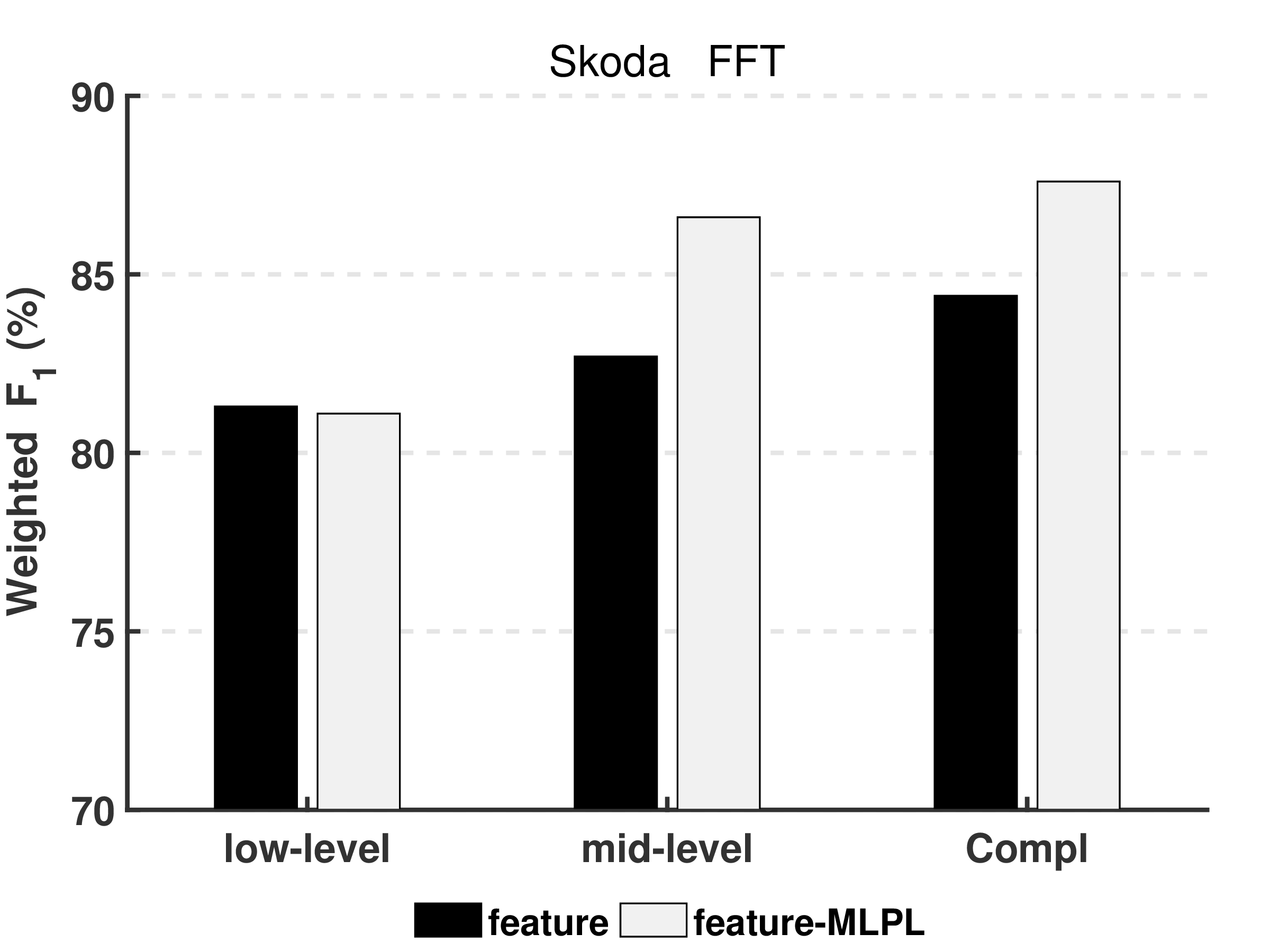}
		\end{minipage}
		\hfill
		\begin{minipage}[b]{.32\textwidth}   
	 	 \includegraphics[width=1\textwidth]{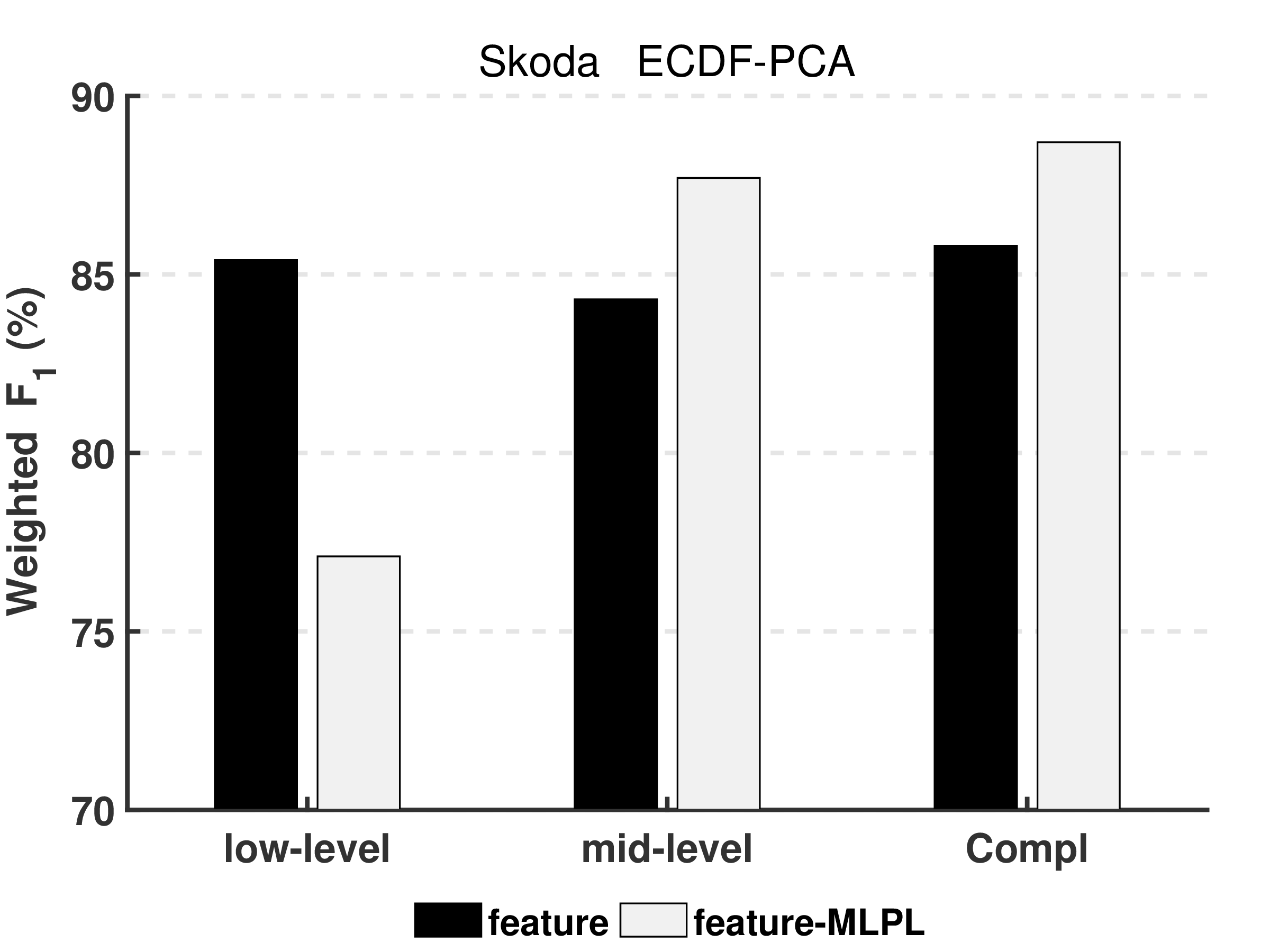}
		\end{minipage}
		\vfill
		\hfill

		\begin{minipage}[b]{.32\textwidth}  
	 	 \includegraphics[width=1\textwidth]{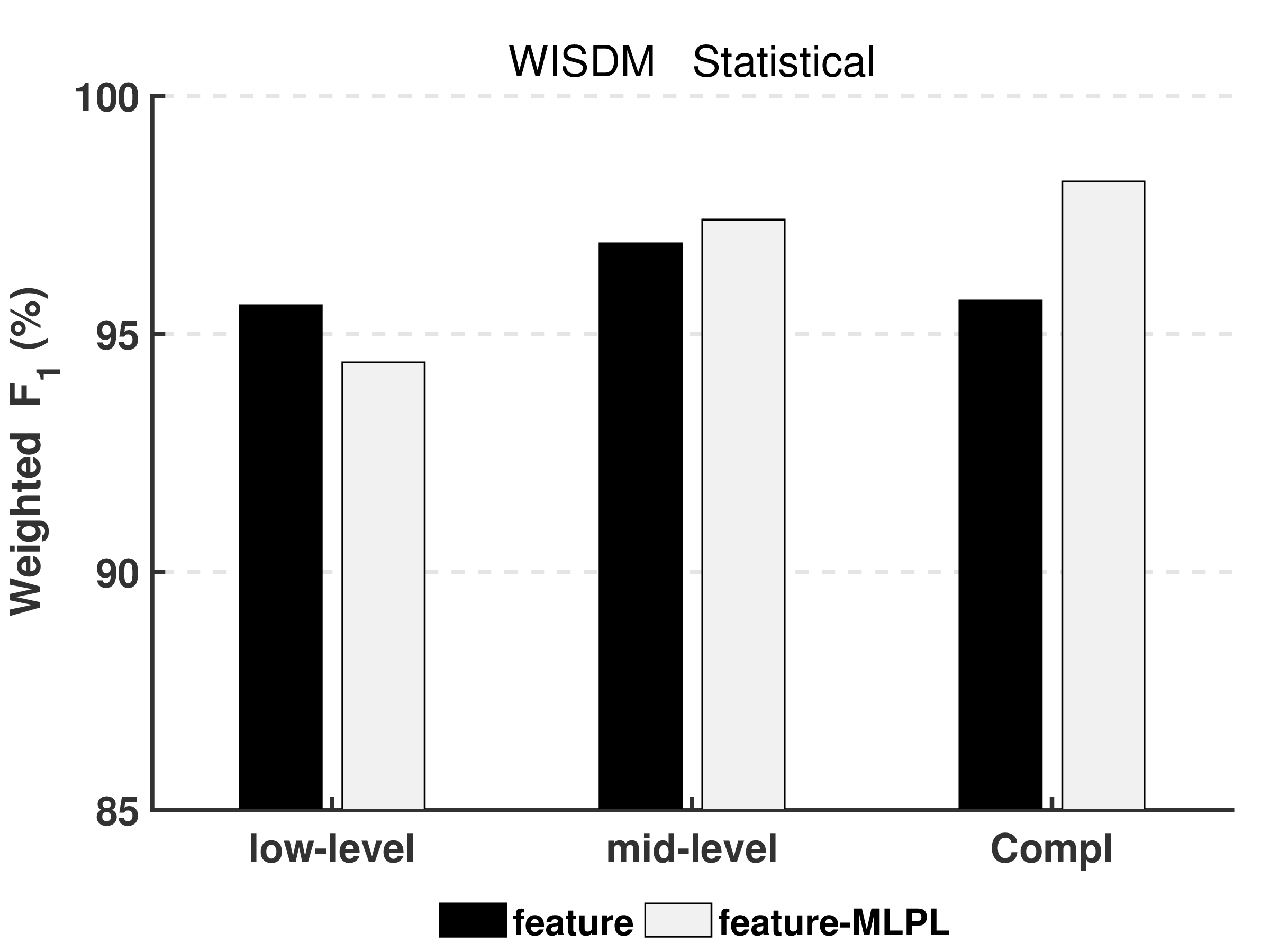}
		\end{minipage}
		\hfill
		\begin{minipage}[b]{.32\textwidth}  
	 	 \includegraphics[width=1\textwidth]{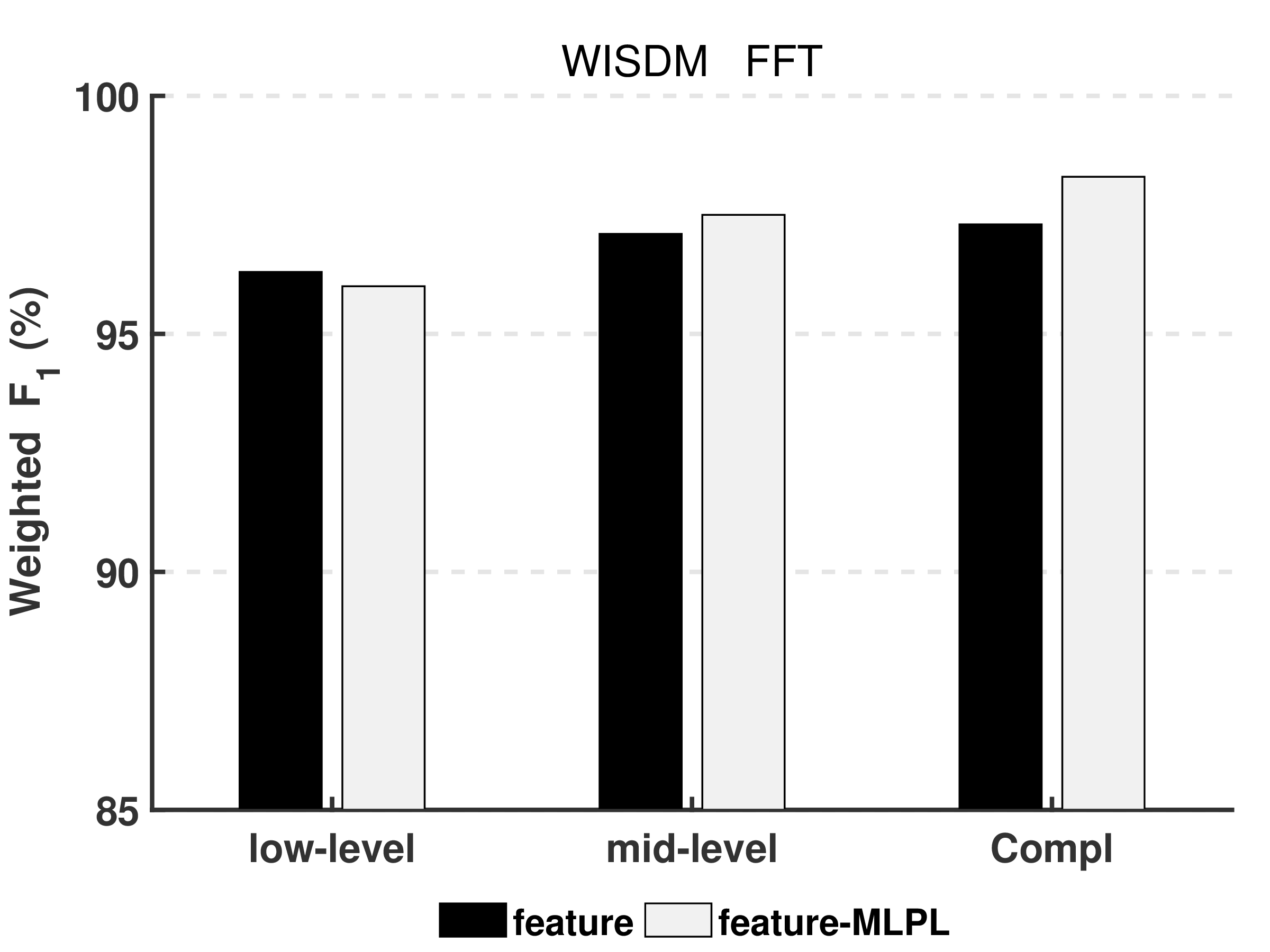}
		\end{minipage}
		\hfill
		\begin{minipage}[b]{.32\textwidth} 
	 	 \includegraphics[width=1\textwidth]{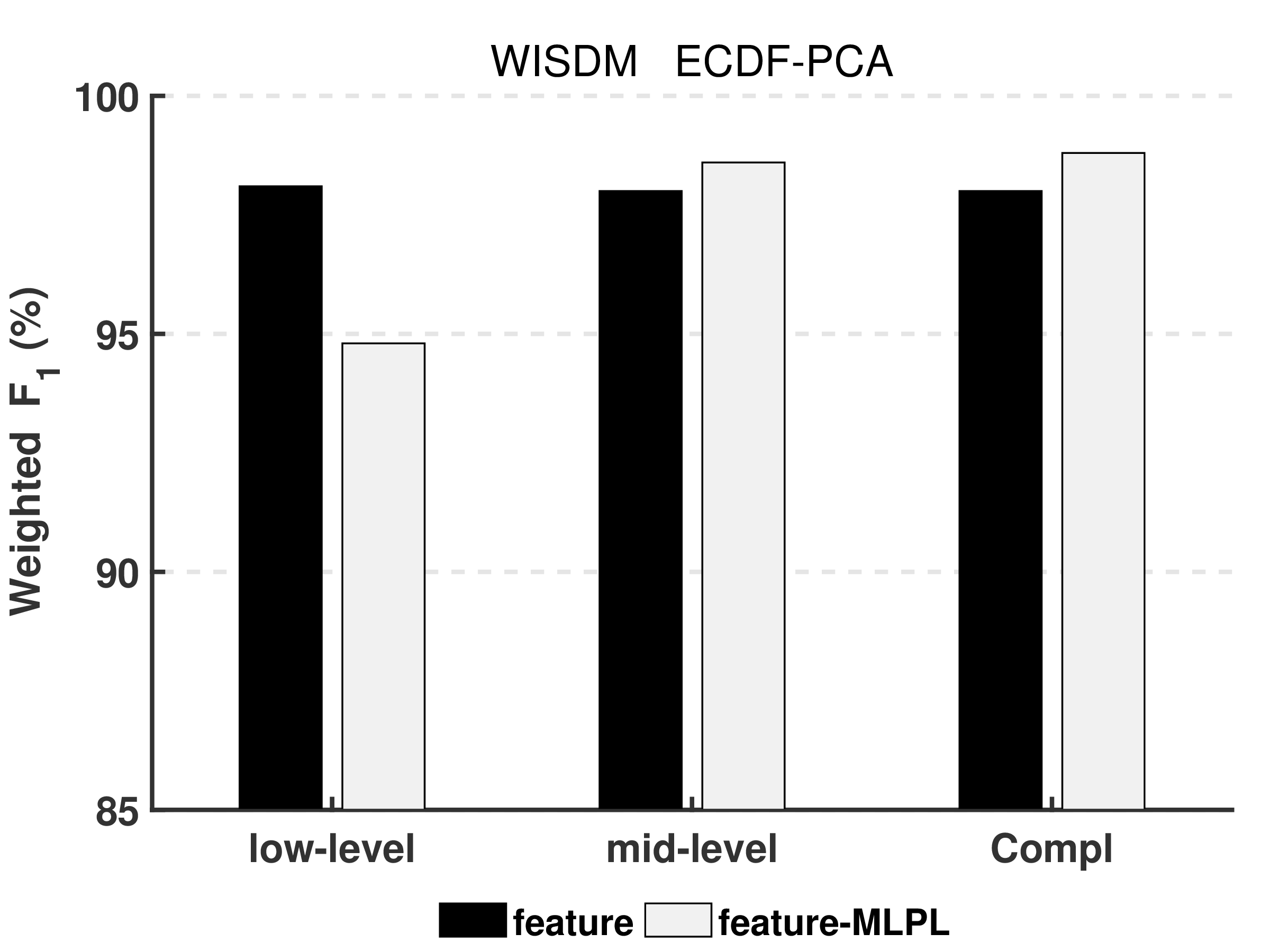}
		\end{minipage}
		\vfill
		\hfill

		\begin{minipage}[b]{.32\textwidth}  
	 	 \includegraphics[width=1\textwidth]{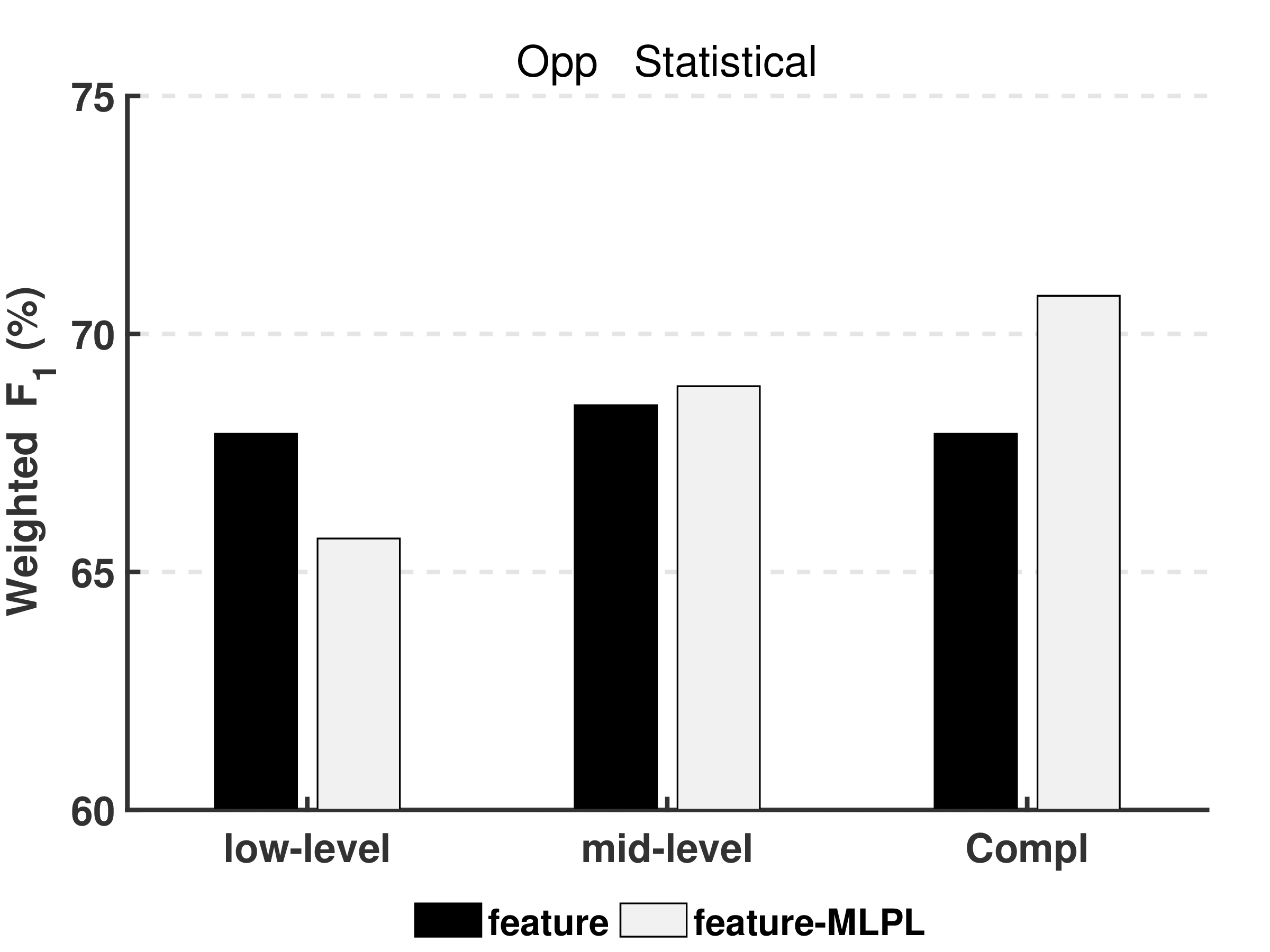}
		\end{minipage}	
		\hfill
		\begin{minipage}[b]{.32\textwidth} 
	 	 \includegraphics[width=1\textwidth]{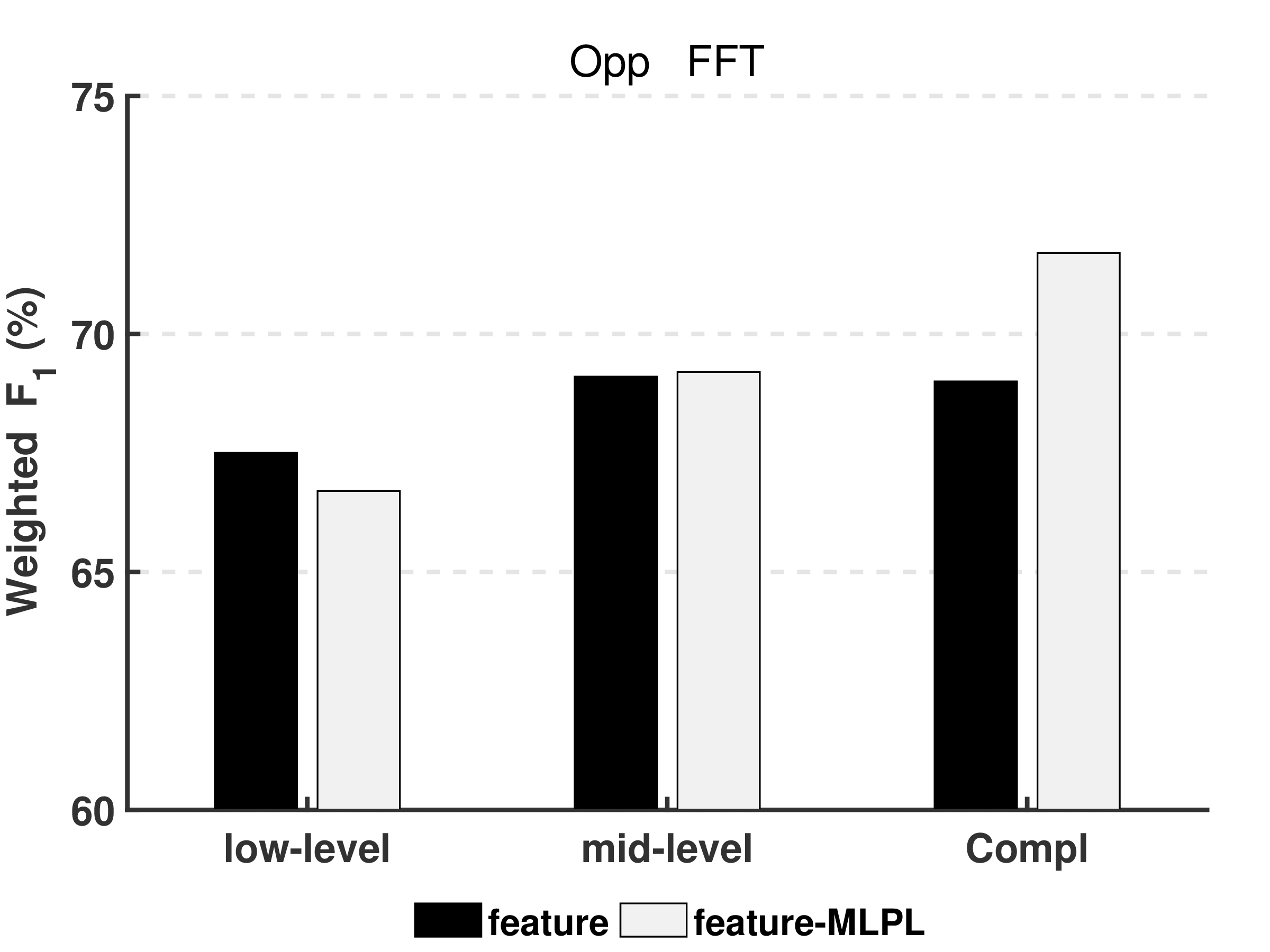}
		\end{minipage}
		\hfill
		\begin{minipage}[b]{.32\textwidth}   
	 	 \includegraphics[width=1\textwidth]{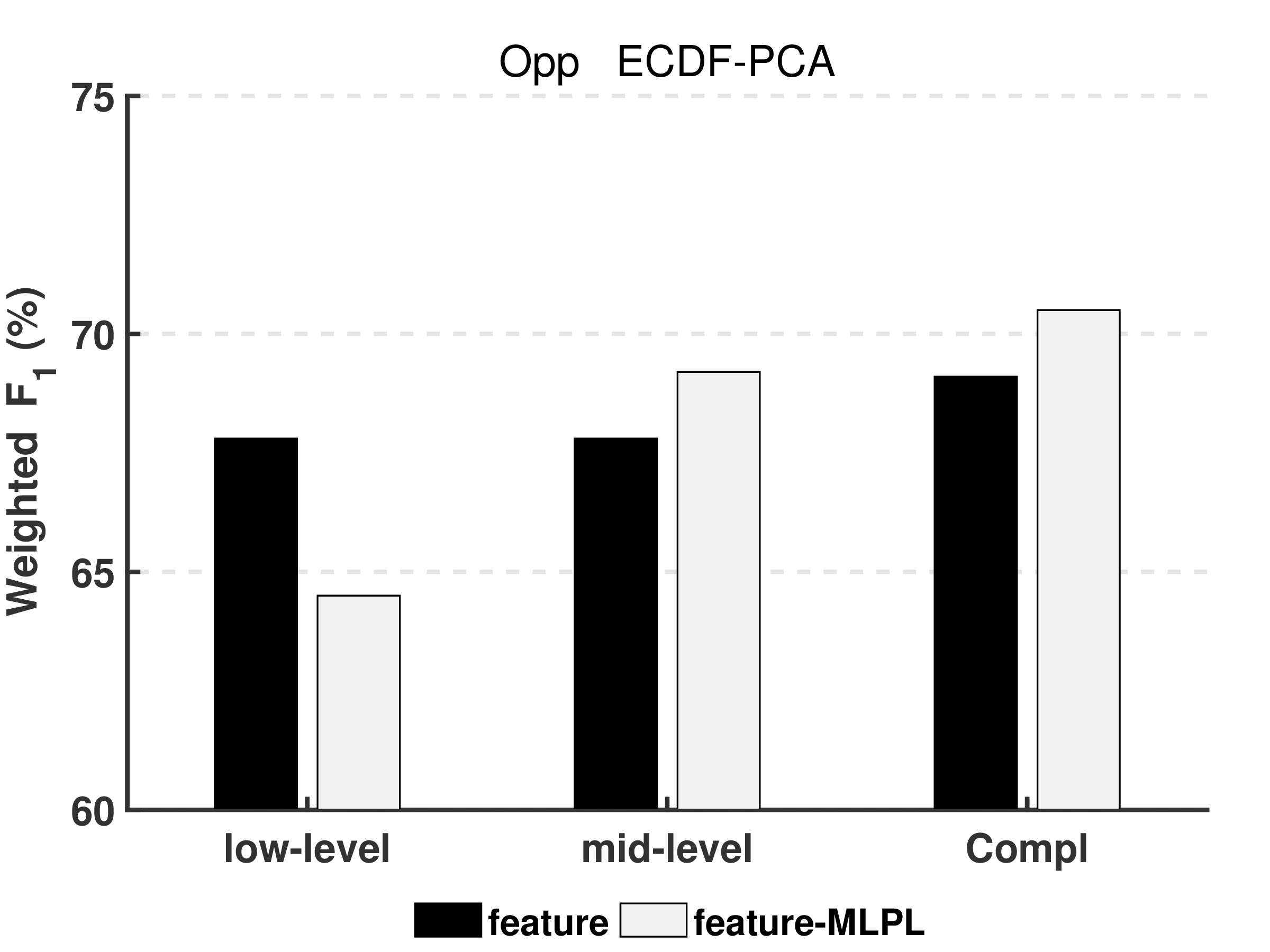}
		\end{minipage}
		
	\caption{Illustrations of the effectiveness of the MLPL process. In each figure, from left to right, black bars show results of low-level feature, mid-level feature and Compl feature; white bars show corresponding results after the MLPL process.}
	\label{fig:fvalue_MLPL_graph}
\end{figure}

In this section, we demonstrate the effectiveness of the Max-margin Latent Pattern Learning (MLPL) method. In particular, we evaluate the MLPL process on three low-level features, their mid-level features and Compl features (the concatenation of low- and mid-level features). Classification results based on KNN are shown in \figurename\ \ref{fig:fvalue_MLPL_graph}.

\figurename\ \ref{fig:fvalue_MLPL_graph} shows that MLPL process achieves remarkable improvements on both mid-level and Compl features. The improvement attributes to their sparse representation which is preferred by the linear SVM solution adopted in MLPL \cite{li2015data}. Compl feature also takes advantage of the complementary property which is further discussed in Section~\ref{sec:lcp}. Results also show that MLCF, which is generated by implementing MLPL on the Compl feature, obtains the highest performance within group comparison on all three datasets. By contrast, performances of low-level features through MLPL degrade, which is partly due to the low-feature being poorly linearly separable. MLPL is designed to explore distinctive distributions in the feature space. Directly implementing MLPL on the low-level feature is not efficient considering that its low dimensional description is not preferred by the linear classifier inside MLPL. In this condition, MLPL would fail to yield strict boundaries among latent classes. 

\subsubsection{Linear complementary property}
\label{sec:lcp}
\begin{figure}[H]
	\centering
	
	 \includegraphics[width=.32\textwidth]{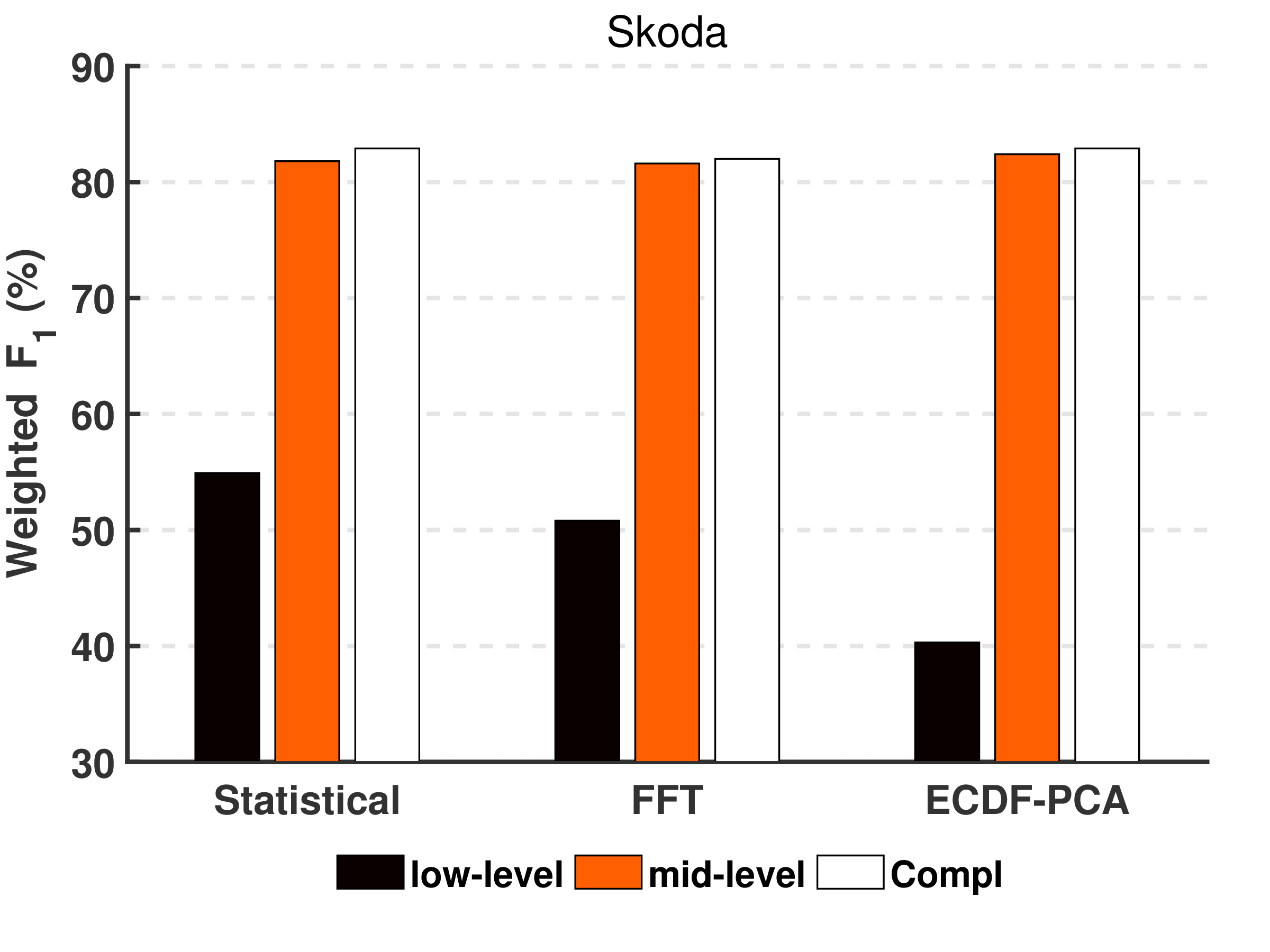}
	 \includegraphics[width=.32\textwidth]{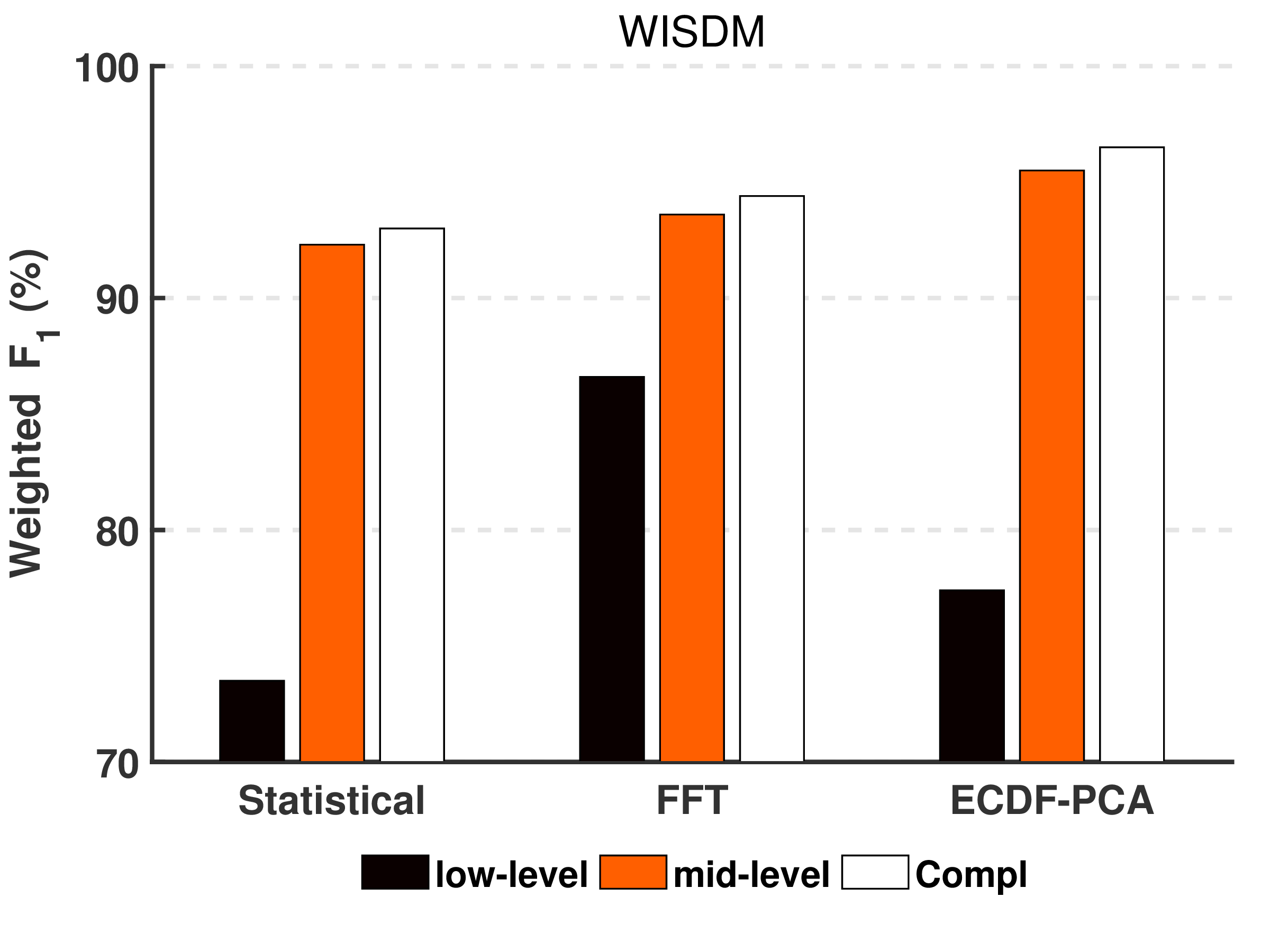}
 	\includegraphics[width=.32\textwidth]{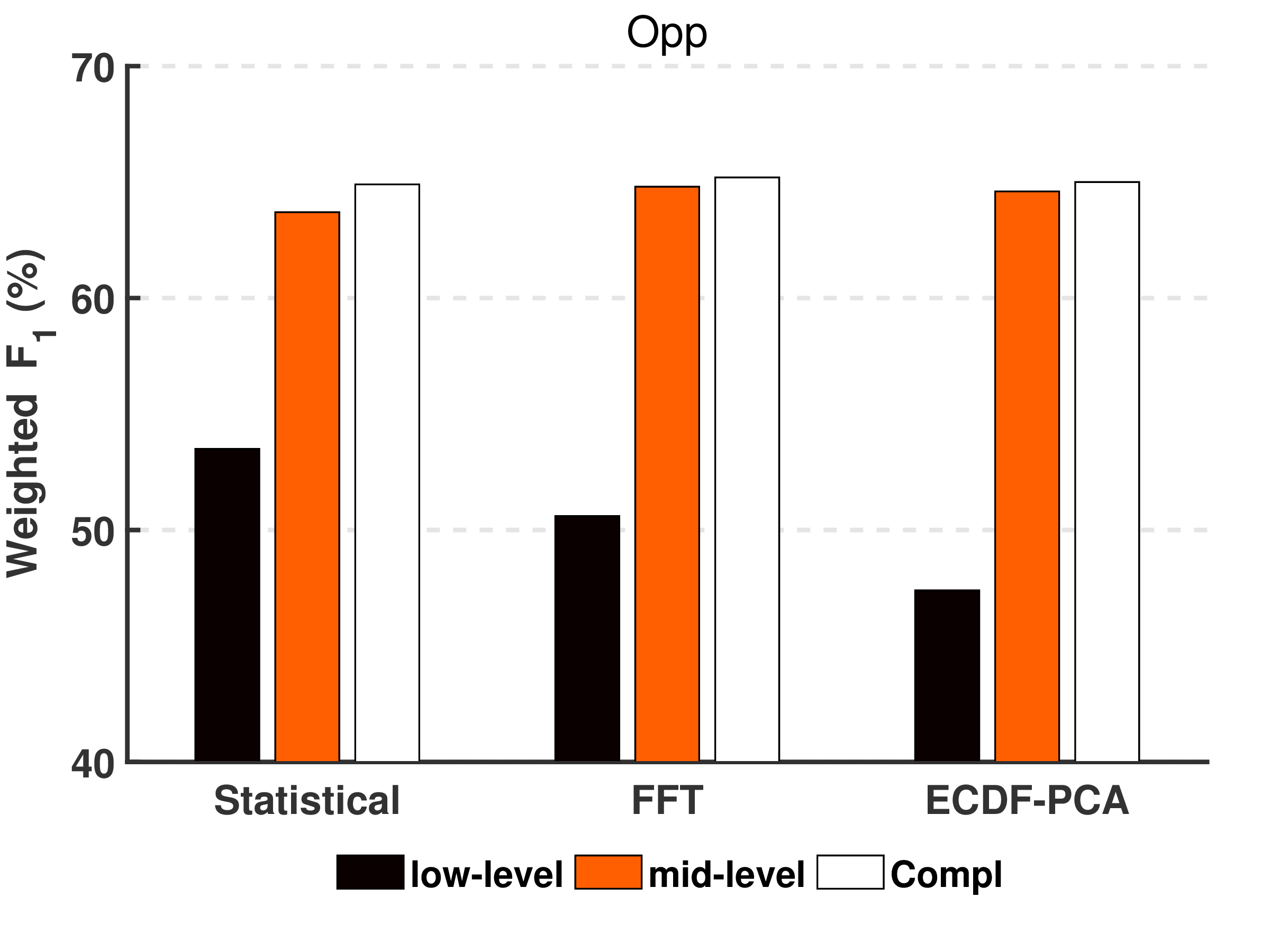}
	\caption{Comparison of Compl feature towards low- and mid-level features using linear SVM.}
	\label{fig:fvalue_complementarity_linear}
\end{figure}

In this section, we demonstrate how complementary property of low-level and mid-level features improves the performance. In particular, experiments are conducted to present intuitive comparisons among low-, mid- and Compl features using linear SVM. Results are shown in Fig. \ref{fig:fvalue_complementarity_linear}.

\figurename\ \ref{fig:fvalue_complementarity_linear} shows that Compl feature performs relatively better than low-level and mid-level features separately with linear classifier. As the mid-level feature is based on statistical description of motion primitives, it contains information from a structural perspective. By contrast, the low-level feature extracted from the raw data can capture detailed information like statistics property (Statistics values), frequency property (FFT coefficients) and amplitude distribution (ECDF-PCA). Therefore, low- and mid-level features are designed to describe the signal from different perspectives \cite{huỳnh2007scalable}. Besides, SVM with linear kernel shows advantages in this condition. This is because in the modeling process, feature weights have inherent characters for feature selection. This property is especially exploited and enhanced in MLPL considering that clusters of linear classifiers are involved. Furthermore, latent classes are initialized by K-means in MLPL, thus clusters are likely to be discriminative in distribution. All those factors contribute to gaining better performances when implementing MLPL on Compl feature. 

\subsubsection{Intra- and inter-personal experiments}

Intra-personal and inter-personal experiments are also conducted to evaluate our framework in daily life scenes. We perform intra-personal tests on Skoda and inter-personal tests on WISDM. In intra-personal experiments, since data in Skoda is recorded by one subject, we divide each class into 6 parts in time sequences and perform 6-fold cross validation. Results obtained by KNN and SVM are shown in Table \ref{table:intra}. In inter-personal experiments, we randomly divide 36 subjects into 10 groups. 10-fold cross validation is conducted on this dataset. The results are shown in Table \ref{table:inter}. 

\begin{table*}[!htbp]
	\caption{\label{table:intra} \newline Comparison (weighted $F_1$ score) among low-level, mid-level features and corresponding MLCF in intra-personal experiments on Skoda using KNN and SVM.} 
	\centering
	\small
	\scalebox{1}{\textbf{}}
	\scalebox{1}{
	\begin{tabular}{cccccccccc}
		\hline
		\multirow{2}{*}{Feature} &
		\multicolumn{3}{c}{KNN} & 
		\multicolumn{3}{c}{SVM}\\
		\cline{2-7}
		& low-level & mid-level & MLCF & low-level & mid-level & MLCF\\
		\hline
		Statistical & 80.4 & 77.7 & \textbf{84.4} & 53.4 & 76.8 & \textbf{78.9}\\
		FFT & 77.9 & 70.6 & \textbf{82.5} & 50.2 & 75.3 & \textbf{76.9}\\
		ECDF-PCA & 81.5 & 75.0 & \textbf{83.0} & 40.0 & 76.1 & \textbf{77.6}\\
		\hline
	\end{tabular}}
\end{table*}

\begin{table*}[!htbp]
	\caption{\label{table:inter} \newline Comparison (weighted $F_1$ score) among low-level, mid-level features and corresponding MLCF in inter-personal experiments on WISDM using KNN and SVM.} 
	\centering
	\small
	\scalebox{1}{\textbf{}}\\
	\scalebox{1}{
	\begin{tabular}{cccccccccc}
		\hline
		\multirow{2}{*}{Feature} &
		\multicolumn{3}{c}{KNN} & 
		\multicolumn{3}{c}{SVM} \\
		\cline{2-7}
		& low-level & mid-level & MLCF & low-level & mid-level & MLCF\\
		\hline
		Statistical & 78.0 & 76.8 & \textbf{80.7} & 71.0 & 79.3 & \textbf{81.4}\\
		FFT & \textbf{83.7} & 68.3 & \textbf{83.6} & 85.1 & 77.0 & \textbf{86.5}\\
		ECDF-PCA & 77.0 & 74.5 & \textbf{79.4} & 73.8 & 81.0 & \textbf{82.8}\\
		\hline
	\end{tabular}}
\end{table*}

MLCF with SVM and KNN generally gain the best performance in both intra-personal and inter-personal test, which can be observed from Table \ref{table:intra} and Table \ref{table:inter}, confirming the robustness of our feature learning framework. In intra-personal tests, MLCF yields better performance in KNN than in SVM, suggesting that similarity strategy adopted by KNN is more suitable for MLCF in inter-personal tasks. Besides, high performance MLCF obtained can be contributed to the fact that action patterns of specific subject have been learned during training and thus can be decently represented in testing. By contrast, in inter-personal experiments, MLCF performs better in SVM than in KNN, owing to SVM's flexibility \cite{belousov2002flexible} in dealing with varieties among subjects.

\subsubsection{Complexity}
We also estimate the overall complexity. Approximately, given fixed-length sampling series, in the first (low-level) stage, the complexity depends on types of low-level features extracted. In the second (mid-level) stage, the complexity is proportional to the product of the size of dictionary and frame. In the third (high-level) stage, the complexity is proportional to the size of latent classes. A typical training and testing time in our experiment at Sokda using FFT as the low feature and Liblinear as classifier is 16 min in training 16,500 samples and 1 minutes in testing 5,500 samples (i7-7700HQ, 2.80 GHZ, 8GB).

\subsection{Sensitivity of parameters}
This section elucidates the evaluation of the variable sensitivity in our framework, including the number of latent classes in the MLPL phase and the dictionary size in the mid-level feature learning phase.

\begin{figure}[H]
	\centering

	 \includegraphics[width=.32\textwidth]{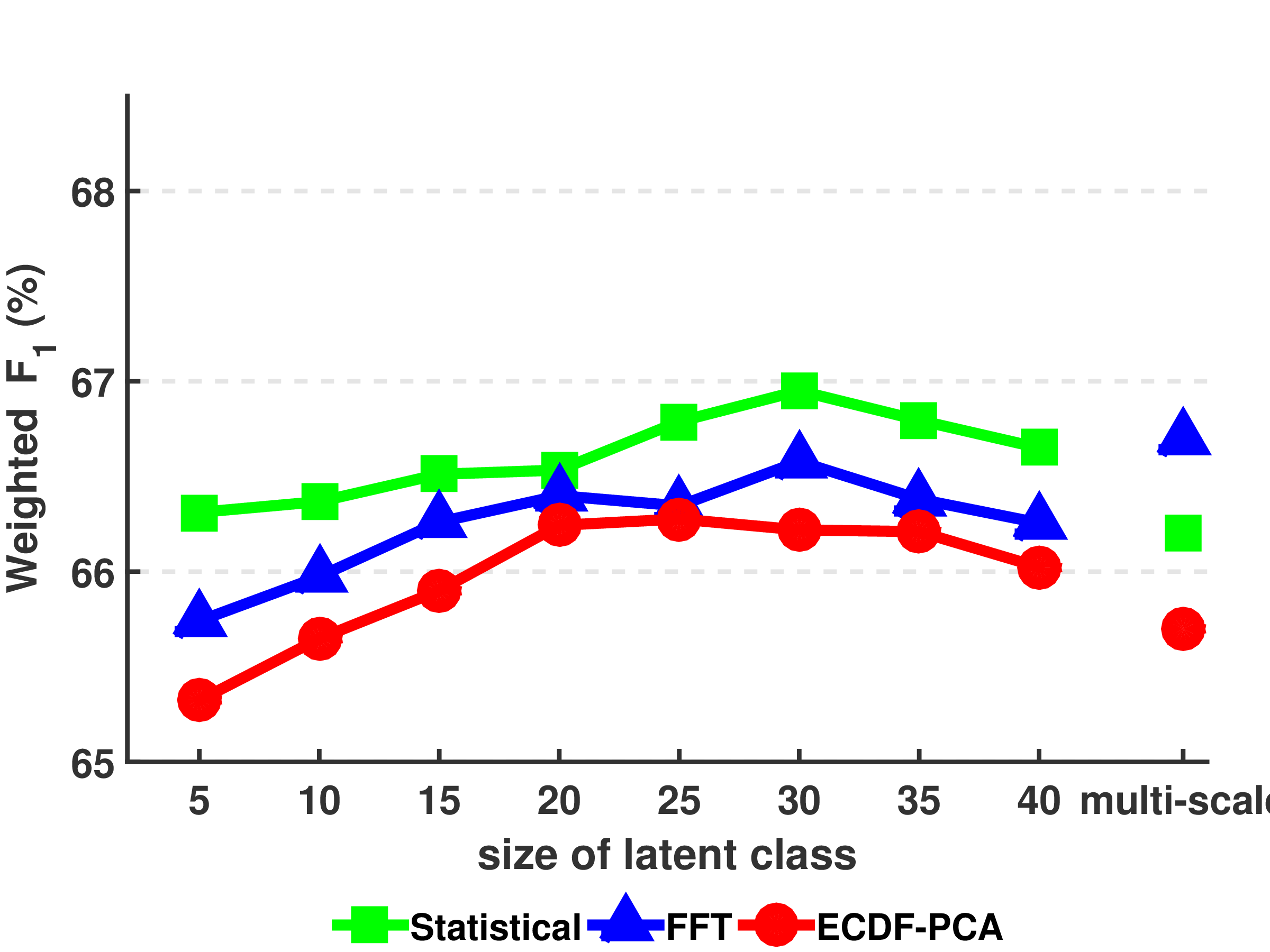}
	 \includegraphics[width=.32\textwidth]{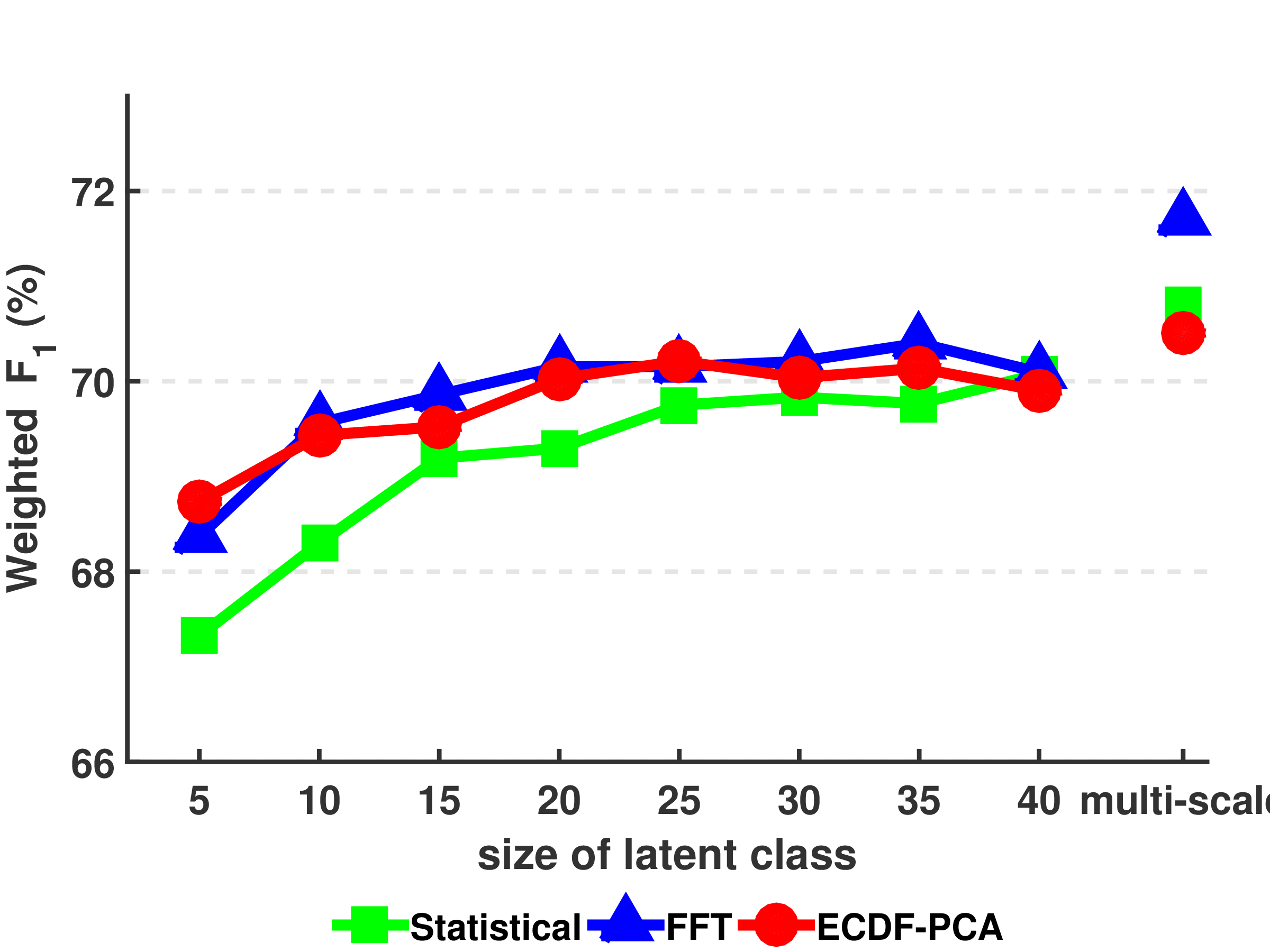}
	 \includegraphics[width=.32\textwidth]{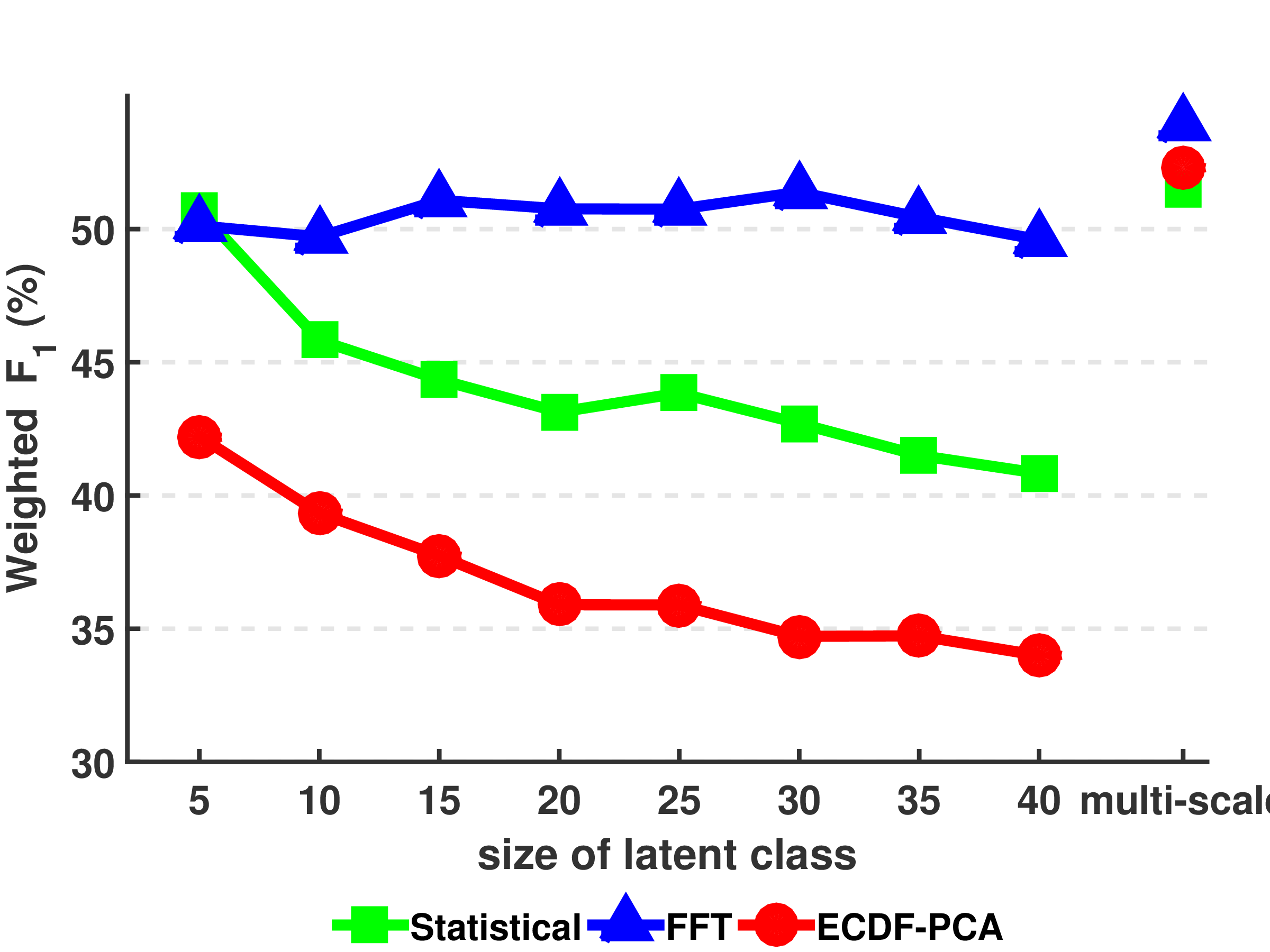}
	\caption{The effect of the latent class size on MLCF. From left to right, classification results of MLCF are obtained by SVM, KNN and NCC, respectively. The `multi-scale' refers to two scales of 5 and 10.}
	\label{fig:latent classes on Opp}
\end{figure}

\subsubsection{Size of latent classes}

We evaluate the sensitivity of the latent class size in MLPL on the Opp dataset, the results of which are shown in \figurename\ \ref{fig:latent classes on Opp}. Latent class sizes are ranging from 5 to 40 and the multi-scale is of 5 and 10. Compared with the single scale, the multi-scale achieves generally the best performance, showing the robustness and effectiveness of learning latent patterns from various semantic levels.
From \figurename\ \ref{fig:latent classes on Opp}, it can also be observed that the tendency of the weighted $F_1$ score is inconsistent according to different classifiers. With SVM, the increasing number of latent classes has positive influence before the size reaches around 30 and after that, the performance slightly degrades. The increase occurs under the circumstance that the specific class is clustered into proper number of groups which are linearly separable in feature space while the degradation can be caused by the case that latent classes are too particular and thus over-represented. Specifically, during MLPL, original properly divided latent classes would be further clustered into small clusters, which are not linearly separable. As MLPL outputs the description of the confidence score of each latent class, over clustering would lead to confusion. The max-margin strategy SVM adopts would be sensitive in this confused description. In terms of KNN, the performance is positively related with increment of the latent class's size, suggesting K-nearest similarity strategy is resistant to over-clustered cases.

\subsubsection{Size of dictionary}

We also change the dictionary size in mid-level feature learning phase to evaluate its influences on Compl feature and MLCF. The experiment is conducted on the Skoda dataset with the size of the motion-primitive dictionary ranging from 100 to 800. 

\begin{figure}[H]
	\centering
	 \includegraphics[width=.32\textwidth]{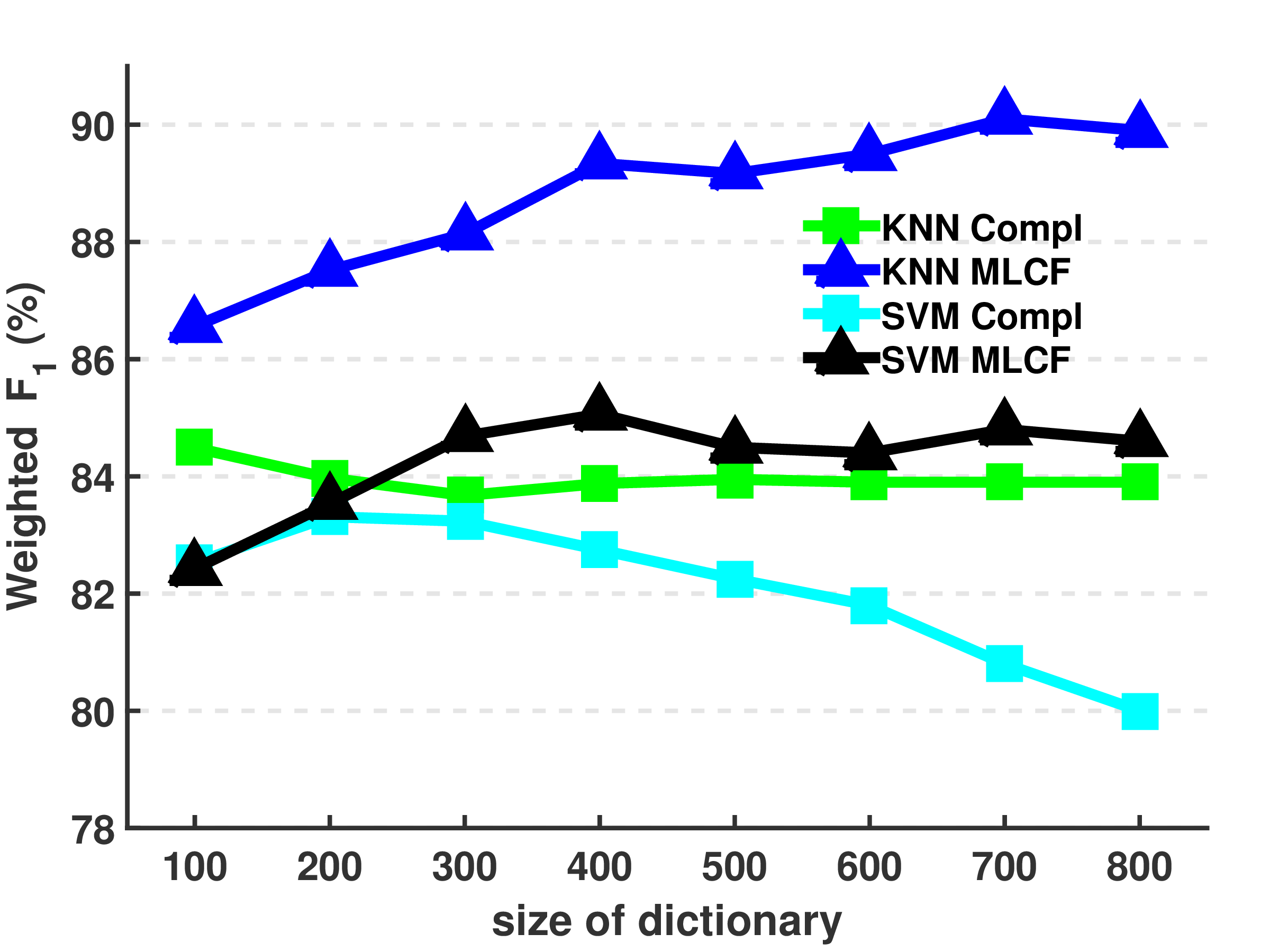}
	 \includegraphics[width=.32\textwidth]{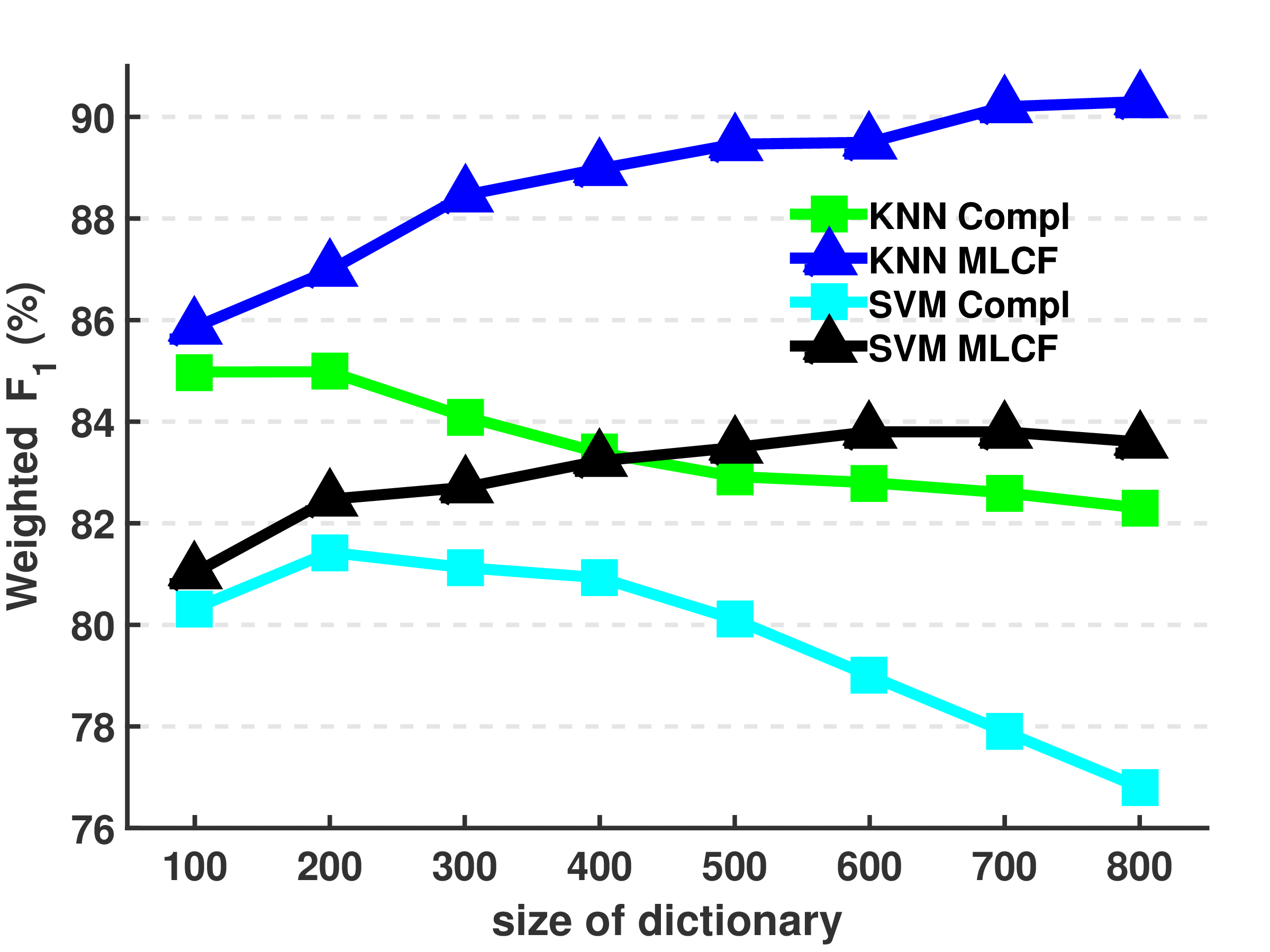}
	 \includegraphics[width=.32\textwidth]{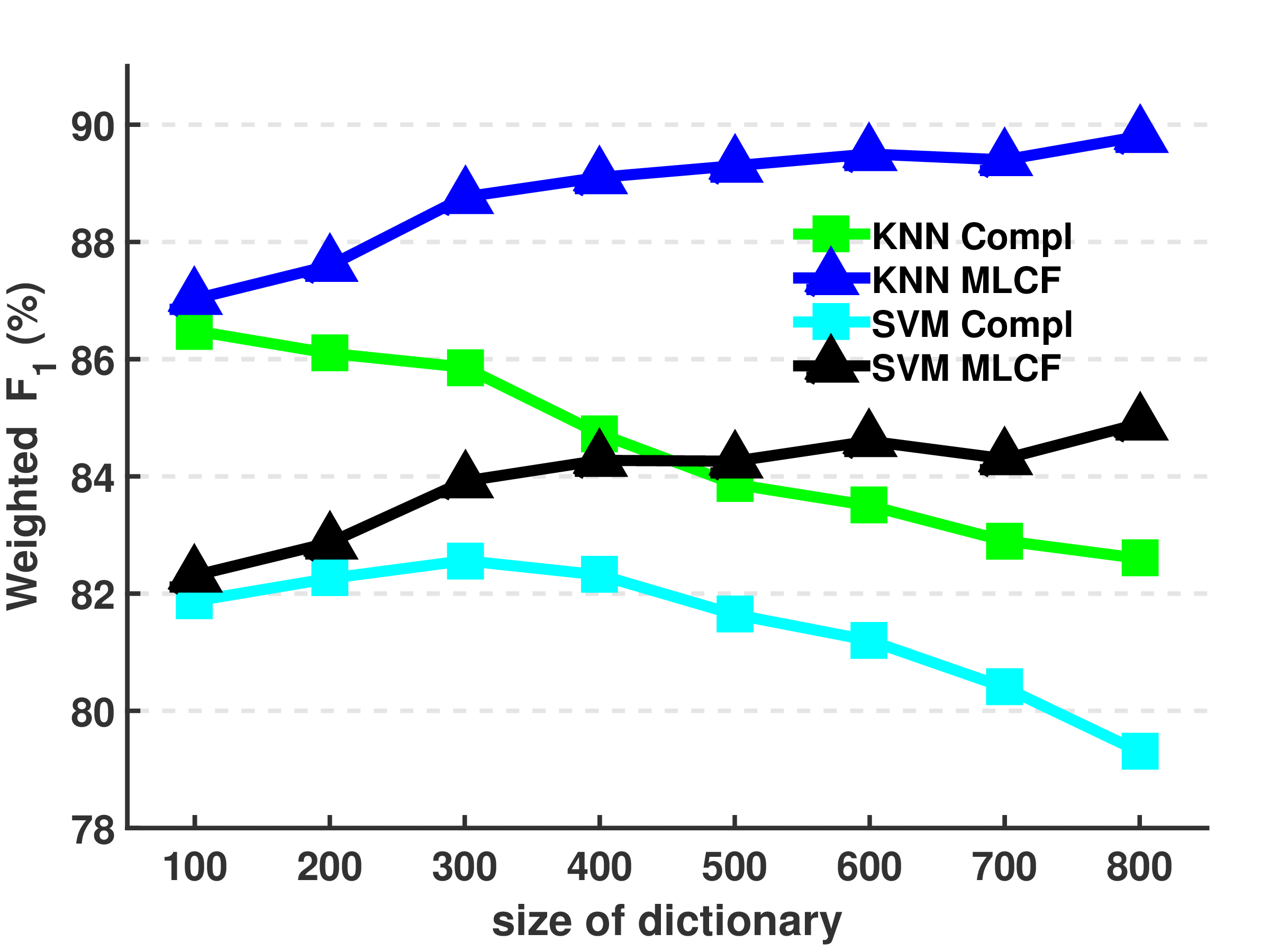}
	\caption{The influence of the motion-primitive dictionary size on Compl feature and MLCF. From left to right, Compl feature and MLCF are derived from statistical values, FFT coefficients and ECDF-PCA, respectively.}	
	\label{fig:size_of_dictionary_skoda}
\end{figure}

As shown in \figurename\ \ref{fig:size_of_dictionary_skoda}, performances of MLCF positively correlate with the dictionary size in both SVM and KNN while performances of Compl feature increase and then slightly degrade in SVM but continuously degrade in KNN. Although the increase in dictionary size involves more detailed motion-primitive description and the statistical process makes feature robust to noise, larger size may lead to excessively redundant representation, especially when features are concatenated in three channels. Under this condition, when Compl features are directly fed into classifiers, data around 22,000 frames are not sufficient in training models. In contrast, MLCF yields better results when the performance of the Compl feature turns to degrade.

\section{Conclusion}

In this paper, we present the MLCFL framework for signal processing in HAR. 
The framework consists of three parts, obtaining low-level, mid-level and high-level features separately. 
The low-level feature captures property of raw signals. 
The mid-level feature achieves component-based representation through hard coding process and occurrence statistics. 
At high level, the latent semantic learning method MLPL is proposed to mine latent action patterns from concatenation of low- and mid-level features, during which the semantic representation can be achieved. 
Our framework achieves the state-of-the-art performances, 88.7\%, 98.8\% and 72.6\% (weighted $F_1$ score) respectively, on Skoda, WISDM and OPP datasets. Given that the MLPL method has the ability of discovering various patterns inside the specific class, it is possible to apply this framework in more challengeable scenarios, like tasks without full annoatations. So a potential improvement of our future work is to merge instance selection processing into current framework in order to deal with wealky learning problems.

\appendix

\section*{Author contributions}
Designed the program and wrote the code: YX ZS XZ SD YG. Provided the data: YX YG. Performed the experiments: ZS XZ SD. Analyzed the data: YG SD YW. Wrote the paper:YX ZS YW YG XZ SD YF EC.


\section*{Acknowledgements}
This work is supported by Microsoft Research under eHealth program, Beijing National Science Foundation in China under Grant 4152033, Beijing Young Talent Project in China, the Fundamental Research Funds for the Central Universities of China under Grant SKLSDE-2015ZX-27 from the State Key Laboratory of Software Development Environment in Beihang University in China.



{
\small 
\begin{spacing}{1}
 \bibliography{bibfile}
 \bibliographystyle{elsarticle-num}
\end{spacing}

}





\end{document}